\documentclass[runningheads]{llncs}
\usepackage[rgb]{xcolor}
\usepackage{pgf}

\usepackage[utf8]{inputenc}
\usepackage{pgfplots}
\usepackage{amsfonts}
\usepackage{amsmath}
\usepackage{amstext}
\usepackage{amssymb}
\usepackage{float}
\usepackage{xspace}
\usepackage[font=footnotesize]{subfig}
\usepackage{multirow}
\usepgfplotslibrary{groupplots}
\pgfplotsset{compat=newest}
\usepackage{shellesc}
\usepackage{booktabs}
\usepackage[english]{babel}

\usepackage{url}
\usepackage{placeins}
\usepackage[ruled,noend]{algorithm2e}
\usepackage[font=small,labelfont=bf,tableposition=top]{caption}

\usepackage{xr-hyper}

\makeatletter
\newcommand*{\addFileDependency}[1]{
  \typeout{(#1)}
  \@addtofilelist{#1}
  \IfFileExists{#1}{}{\typeout{No file #1.}}
}
\makeatother

\newcommand*{\myexternaldocument}[1]{%
    \externaldocument{#1}%
    \addFileDependency{#1.tex}%
    \addFileDependency{#1.aux}%
}

\myexternaldocument{appendix_new}

\usepackage[breaklinks]{hyperref}
\usepackage{url}

\usepackage{placeins}
\usepackage{cleveref}
\crefname{assumption}{Assumption}{assumptions}
\DeclareCaptionLabelFormat{andtable}{#1~#2  \&  \tablename~\thetable}

\addto\extrasenglish{%

}

\begin{document}

\definecolor{mygreen}{rgb}{0.2, 0.7, 0.2}
\definecolor{myorange}{rgb}{0.9, 0.5, 0.0}
\definecolor{mypurple}{rgb}{0.5, 0, 0.5}

\newcommand\noteRC[1]{\textcolor{blue}{RC - #1}}
\newcommand\notePM[1]{\textcolor{green}{PM - #1}}
\newcommand\noteMF[1]{\textcolor{mypurple}{MF - #1}}
\newcommand\noteGF[1]{\textcolor{red}{GF - #1}}

\newcommand{\dashlist}{\renewcommand\labelitemi{--}}
\newcommand{\argmin}{\operatornamewithlimits{argmin}}
\newcommand{\argmax}{\operatornamewithlimits{argmax}}
\newcommand{\infop}{\operatornamewithlimits{inf}}
\newcommand{\minop}{\operatornamewithlimits{min}}
\newcommand{\xvect}{\mathbf{x}}
\newcommand{\yvect}{\mathbf{y}}
\newcommand{\uvect}{\mathbf{u}}
\newcommand{\mvect}{\mathbf{m}}
\newcommand{\fx}{f\left( \xvect \right)}
\newcommand{\fy}{f\left ( \yvect \right )}
\newcommand{\Rd}{\mathbb{R}^{d}}
\newcommand{\gvect}{\mathbf{g}}
\newcommand{\stgrad}{\gvect}
\newcommand{\xt}{\xvect_{t}}
\newcommand{\xl}{\xvect_{l}}
\newcommand{\xtp}{\xvect_{t+1}}
\newcommand{\mt}{\mvect_{t}}
\newcommand{\mtp}{\mvect_{t+1}}
\newcommand{\etat}{\eta_{t}}
\newcommand{\gradfi}{\nabla f \left ( \xt,i \right )}
\newcommand{\gradfitau}{\nabla f \left ( \xvect_{\taut},i \right )}
\newcommand{\gradfl}{\nabla f \left ( \xl \right )}
\newcommand{\gradf}{\nabla f \left ( \xt \right )}
\newcommand{\mbstgrad}{{\gvect} \left ( \xvect_{t},\xi_{t}  \right )}
\newcommand{\taut}{\tau _{t}}
\newcommand{\avtau}{\bar{\tau}}
\newcommand{\ambstgrad}{{\gvect} \left ( \xvect_{\taut},\xi_{t}  \right )}
\newcommand{\phik}{\Phi _{k}}
\newcommand{\fxt}{f\left ( \xt \right )}
\newcommand{\compresgrad}{ \phik \left [ \ambstgrad \right ] }
\newcommand{\compresf}{ \phik \left [ \nabla f \left ( \xt \right ) \right ] }
\newcommand{\gradftau}{\nabla f \left ( \xvect_{\taut} \right )}
\newcommand{\mut}{\mu_{t}}
\newcommand{\egrad}{\mathbb{E} \left[ \left \| \gradf \right \|^{2} \right ]}
\newcommand{\egradl}{\mathbb{E} \left[ \left \| \gradfl \right \|^{2} \right ]}
\newcommand{\sigmasq}{\sigma^{2}}
\newcommand{\ef}{\mathbb{E}\left [ f \left ( \xt \right ) \right ]}
\newcommand{\sumT}{\sum_{t=0}^{T-1}}

\newtheorem{assumption}{Assumption}

\newcommand{\name}[1]{{\textsc{#1}}\xspace}
\newcommand{\vanilla}{\name{ASGD}}
\newcommand{\kardam}{\name{Kardam}}
\newcommand{\local}{\name{LocalSGD}}
\newcommand{\ssgd}{\name{$\phi$SGD}}
\newcommand{\ssgdm}{\name{$\phi$MemSGD}}

\newcommand{\cnn}{\name{Cnn}}
\newcommand{\mnist}{\name{Mnist}}
\newcommand{\cifar}{\name{Cifar10}}
\newcommand{\lenet}{\name{LeNet}}
\newcommand{\alexnet}{\name{AlexNet}}
\newcommand{\resnet}{\name{ResNet-56}}

\title{Convergence Analysis of Sparsified Asynchronous SGD}
%
%
 \author{Rosa Candela \and
 Giulio Franzese \and
 Maurizio Filippone \and Pietro Michiardi}

\authorrunning{R. Candela et al.}
%
\institute{}
\maketitle              
\begin{abstract}
        Large scale machine learning is increasingly relying on distributed optimization, whereby several machines contribute to the training process of a statistical model.
In this work we study the performance of asynchronous, distributed settings, when applying sparsification, a technique used to reduce communication overheads.
In particular, for the first time in an asynchronous, non-convex setting, we theoretically prove that, in presence of staleness, sparsification does not harm SGD performance: the ergodic convergence rate matches the known result of standard SGD, that is $\mathcal{O} \left( 1/\sqrt{T} \right)$.
We also carry out an empirical study to complement our theory, and confirm that the effects of sparsification on the convergence rate are negligible, when compared to ``vanilla’’ SGD, even in the challenging scenario of an asynchronous, distributed system. 

\end{abstract}

    \newlength\figureheight
    \newlength\figurewidth

    \section{Introduction}
    \label{sec:introduction}
The analysis of Stochastic Gradient Descent (SGD) \cite{RobbinsMonro:1951} and its variants has received a lot of attention
recently, due to its popularity as an optimization algorithm in machine learning; see \cite{Bottou:2018} for
an overview.
SGD addresses the computational bottleneck of gradient descent by relying on stochastic gradients, which are 
cheaper to compute than full gradients.
SGD trades a larger number of iterations to converge for a cheaper cost per iteration.
The \emph{mini-batch} variant of SGD allows one to control the number and the cost per iteration, making it the preferred
choice for optimization in deep learning \cite{bottou2010large,Bottou:2018}.

We consider the problem of optimizing the $d$-dimensional parameter vector $\xvect \in \mathbb{R}^{d}$
of a model and its associated finite-sum \emph{non-convex}
loss function $\fx = \frac{1}{n}\sum_{i=1}^{n}f\left ( \xvect,i \right )$, where
$f\left ( \xvect,i \right ) \text{, } i=1,\ldots ,n$ is the loss function for a single training sample $i$.
SGD iterations have the following form:
\[
\xvect_{t+1} = \xt - \etat \stgrad\left ( \xt,i \right ),
\]
where $\xt, \xvect_{t+1} \in \Rd$ are the model iterates, $\etat > 0$ is the learning rate/step size and
$\stgrad \left ( \xt,i \right )= \gradfi $ is a stochastic gradient.

In this work, we are interested in the increasingly popular distributed setting, whereby SGD runs across several machines,
which contribute to the model updates $\xvect_{t+1}$ by computing stochastic gradients of the loss using locally available
training data \cite{NIPS2012_4687,Ho:2013,186212,186214,Jiang:2017}.
The analysis of the convergence behavior of SGD, both in synchronous \cite{45187,Zhang2016ParallelSW,Bottou:2018}
and asynchronous \cite{NIPS2011_4390,NIPS2015_5717,Lian:2015:APS:2969442.2969545,Liu:2015,Bottou:2018} settings
has been widely studied in the literature.
In this work, we focus on the asynchronous setting, which is particularly challenging because
distributed workers might produce gradient updates for a loss computed on \emph{stale} versions of the current model iterates
\cite{Ho:2013,Damaskinos:2018,Lin:2018,pmlr-v84-dutta18a,Dai:2019}.

In this context, communication overheads have been considered as a key issue to address, and a large
number of works have been proposed to mitigate such overheads
\cite{Stich:2018,Yu2019ParallelRS,NIPS2018_7405,NIPS2017_6768,NIPS2017_6749,DBLP:journals/corr/AjiH17,7835789}.
In particular, sparsification methods \cite{NIPS2018_7697,NIPS2018_7837,NIPS2018_7405} have achieved remarkable results,
albeit for synchronous setups.
The key idea is to apply smaller and more efficient gradient updates, by applying a sparsification operator
to the stochastic gradient, which results in updates of size $k \ll d$.

In this work, we fill the gap in the literature and study sparsification methods in \emph{asynchronous settings}.
For the first time, we provide a concise and simple convergence rate analysis when the joint effects of
sparsification and asynchrony are taken into account, and show that sparsified SGD converges at the same rate of standard SGD.
Our empirical analysis of sparsified SGD complements our theory.
We consider several delay distributions and show that, in practice, applying sparsification does not harm SGD performance.
These results carry over when the system scales out, which is a truly desirable property.

\subsection{Related work}
\label{sec:related:work}

The analysis of SGD \cite{RobbinsMonro:1951} and its convergence properties has recently attracted a lot of attention,
especially in the field of machine learning \cite{Bottou:2018,NIPS2011_4316}, where SGD is considered the workhorse
optimization method.
Large scale models and massive datasets have motivated researchers to focus on distributed machine learning, whereby
multiple machines compute stochastic gradients using partitions of the dataset and a parameter server maintains a globally
shared model.

Asynchronous systems \cite{NIPS2011_4390,NIPS2012_4687,186214,186212} provide fast model updates, but the use of
stale parameters might affect convergence speed.
One way to reduce the staleness effect is to give a smaller weight to stale updates.
In \cite{Jiang:2017,Damaskinos:2018} gradient contributions are dampened through a dynamic learning rate.
Stale-synchronous parallel (SSP) models \cite{Ho:2013,Damaskinos:2018} limit instead the maximum staleness,
discarding updates that are too ``old''.
Interestingly, the work in \cite{mitliagkas2016asynchrony}, suggests to view staleness as a form of implicit momentum,
and study, under a simple model, how to adjust explicit, algorithmic momentum to counterbalance the effects of staleness.

Synchronous systems \cite{45187} guarantee higher statistical efficiency, but the presence of stragglers slows down the
learning algorithm.
One solution is provided by the so called local SGD models \cite{Lin:2018,Stich:2018,Yu2019ParallelRS}, which reduce
the synchronization frequency by allowing nodes to compute local model parameters, which are averaged in a global
model update.
A second family of approaches seeks to improve synchronous systems by reducing the cost of communicating gradients
upon every iteration.
Quantization techniques reduce the number of bits to represent the
gradients before communication \cite{seide20141,NIPS2017_6768,NIPS2017_6749},
sparsification methods select a subset of the gradient
components to communicate \cite{DBLP:journals/corr/AjiH17,7835789,strom2015scalable,NIPS2018_7697,NIPS2018_7837,lin2018deep},
and loss-less methods use large mini-batches to
increase the computation-communication ratio \cite{Goyal2017AccurateLM,you2017scaling}.

Our work, along the lines of \cite{Lian:2015:APS:2969442.2969545,NIPS2018_7837}, argues instead that staleness vanishes, asymptotically.
Similarly, recent work \cite{stich2019errorfeedback} uses an elegant analysis technique to study the role of stale gradient updates
and sparsification, albeit their effects are considered in isolation.
In this work, instead, we provide a concise and simple convergence rate analysis for the joint effects of sparsification
and staleness.

\subsection{Contributions}
\label{subsec:contributions}

%

We study finite-sum non-convex optimization of loss functions of the form $\fx: \mathbb{R}^{d} \to \mathbb{R}$, and assume that $f$
is continuously differentiable and bounded below, that $\nabla \fx$ is $L$-Lipschitz smooth, that the variance of
stochastic gradients is bounded, and that the staleness induced by asynchrony is also bounded.
We analyze a mini-batch asynchronous SGD algorithm 
and apply a sparsification operator $\phik \left [ \ambstgrad \right ]$ with $k \ll d$, which can be coupled with
an error correction technique, often called \emph{memory} \cite{NIPS2018_7697}.

We prove ergodic convergence of the gradient of $\fx$, for an appropriately chosen learning rate.
In particular, we focus on memory-less variants, which are simpler to analyze, and show that asynchronous sparsified SGD
converges at the same rate as standard SGD.

In this paper, the main theoretical contribution is as follows. 
Let the sparsification coefficient be $\rho=k/d$.
Then, it holds that:
\[
\minop\limits_{0\leq t\leq T}\egrad\leq \frac{\left(\sumT\left(\frac{\etat^2 L}{2}\sigmasq\right)\right)+\Lambda+C}{\sumT\left(\etat\rho\mu-\frac{\etat^2 L}{2}\right)},
\]
where  $\Lambda = f \left ( \xvect_{0} \right ) - \infop\limits_{\xvect} \fx$ and $C,\mu$ are finite positive constants (whose role will be clarified later). In particular for a suitable constant learning rate $\etat=\eta=\frac{\rho\mu}{L\sqrt{T}}$ we can derive as a corollary that:
\begin{flalign*}
&\minop\limits_{0\leq t\leq T}\egrad\leq\left(\frac{\sigmasq}{2}+\frac{(\Lambda+C)L}{(\rho\mu)^2}\right)\frac{1}{\sqrt{T}},
\end{flalign*}
up to a negligible approximation for large $T$ (details in the supplement).

We define sparsified SGD formally in \Cref{sec:sasgd}, both in its memory and memory-less variants, and outline our
proof for the memory-less case in \Cref{subsec:sketch}.
In \Cref{sec:experiments} we provide an empirical study of the convergence behavior
of the two variants of sparsified SGD, using simple and deep convolutional networks for image classification tasks.
Our experiments show that sparsification does not harm SGD performace, even in the challenging scenario of an asynchronous, distributed system.
Although we do not provide convergence guarantees for sparsified SGD with memory, our empirical results indicate that
error correction dramatically improves the convergence properties of the algorithm.

    \section{Sparsified Asynchronous SGD}
    \label{sec:sasgd}
    In this Section we define two variants of sparsified SGD algorithms, with and without error correction, and
emphasize the role of model staleness induced by the asynchronous setup we consider.

The standard way to scale SGD to multiple computing nodes is via \emph{data-parallelism}: a set of worker
machines have access to the $n$ training samples through a distributed filesystem.
Workers process samples concurrently: each node receives a copy of the parameter vector $\xt$, and computes
stochastic gradients locally.
Then, they send their gradients to a parameter server (PS).
Upon receiving a gradient from a worker, the PS updates the model by producing a new iterate $\xtp$.

Due to asynchrony, a computing node may use a \emph{stale} version of the parameter vector:
a worker may compute the gradient of $f\left( \mathbf{x}_{\tau_t} \right),\, \tau_t \leq t$.
We call $\tau_t$ the \emph{staleness of a gradient update}.
As stated more formally in \Cref{sec:convergence}, in this work we assume \textbf{bounded staleness}, which is
realistic in the setup we consider.
Other works, e.g. that consider Byzantine attackers \cite{Damaskinos:2018}, drop this assumption.\\

\noindent \textbf{Gradient sparsification.} A variety of compression \cite{Bernstein2018CompressionBT,bernstein2018signsgd},
quantization \cite{7835789,NIPS2017_6768} and
sparsification \cite{NIPS2018_7837,NIPS2018_7697} operators have been
considered in the literature.
Here we use sparsification, defined as follows:
\begin{definition}
    \label{def:topk}
    Given a vector $\uvect \in \mathbb{R}^{d}$, a parameter $1 \leq k \leq d$, the operator $\phik(\uvect):
    \mathbb{R}^{d} \to \mathbb{R}^{d}$ is defined as:
    \[
        (\phik(\uvect))_i =
        \begin{cases}
            (\uvect)_{\pi(i)},& \mbox{if } i \leq k, \\
            0,& \mbox{otherwise}
        \end{cases}
    \]
    where $\pi$ is a permutation of the indices $\{1, \ldots, d\}$ such that $(|\uvect|)_{\pi(i)} \geq (|\uvect|)_{\pi(i+1)},
    \forall i \in {1, \cdots, d}$.
\end{definition}

Essentially, $\phik(\cdot)$ sorts vector elements by their magnitude, and keeps only the top-$k$.
A key property of the operator we consider is called the $k$-contraction property \cite{NIPS2018_7697},
which we use in our convergence proofs.

\begin{definition}
    \label{def:kcontraction}
    For a parameter $1 \leq k \leq d$, a $k$-contraction operator $\phik(\uvect):
    \mathbb{R}^{d} \to \mathbb{R}^{d}$ satisfies the following contraction property:
    \[
        \mathbb{E} \left\| \uvect - \phik(\uvect)  \right\|^2 \leq \left( 1 - \frac{k}{d} \right) \|\uvect\|^2.
    \]
\end{definition}
Both the top-$k$ operator we consider, and randomized variants, satisfy the $k$-contraction property
\cite{NIPS2018_7837,NIPS2018_7697}.
Next, we state a Lemma that we will use for our convergence rate results.

\begin{lemma}
    \label{lemma:comp2vect}
    Given a vector $\uvect \in \mathbb{R}^{d}$, a parameter $1 \leq k \leq d$, and the top-$k$ operator $\phik(\uvect):
    \mathbb{R}^{d} \to \mathbb{R}^{d}$ introduced in \Cref{def:topk}, we have that:
    \[
        \|\phik(\uvect)\|^2 \geq \frac{k}{d} \|\uvect\|^2.
    \]
\end{lemma}
The proof of \Cref{lemma:comp2vect} uses the $k$-contraction property in \Cref{def:kcontraction},
as shown in \Cref{sec:comp2vect}.\\

\noindent \textbf{Memory and memory-less sparsified asynchronous SGD.}
We define two variants of sparsified SGD: the first uses sparsified stochastic
gradient updates directly, whereas the second uses an error correction technique which accumulates information suppressed by
sparsification. 
Since we consider an asynchronous, \emph{mini-batch} version of SGD, additional specifications are in order.

\begin{definition}
    Given $n$ training samples, let $\xi_{t}$ be a set of indices sampled uniformly at random from $\{1, \cdots, n\}$, with
    cardinality $|\xi_{t}|$.
    Let $\taut$ be the bounded \emph{staleness} induced by the asynchronous setup, with respect to the current iterate $t$.
    That is, $t-S\leq\taut \leq t$.
    A stale, mini-batch stochastic gradient is defined as:

    \[
        \ambstgrad = \frac{1}{|\xi_{t}|} \sum_{i \in \xi_{t}} \gradfitau.
    \]
\end{definition}

\paragraph{Memory-less sparsified SGD.}
Given the operator $\phik(\cdot)$, the memory-less, asynchronous sparsified SGD algorithm amounts
to the following:
\[
    \label{algo:memory-less}
    \xtp = \xt - \etat \phik\left( \ambstgrad \right),
\]
where $\{\etat\}_{t \geq 0}$ denotes a sequence of learning rates.

\paragraph{Sparsified SGD with memory.}
Given the operator $\phik(\cdot)$ , the asynchronous sparsified SGD with memory algorithm is
defined as:
\begin{align*}
    \label{algo:memory}
    \xtp &= \xt - \etat \phik\left( \mt + \ambstgrad \right), \\
    \mtp &= \mt + \ambstgrad - \phik\left( \mt + \ambstgrad \right),
\end{align*}
where $\{\etat\}_{t \geq 0}$ denotes a sequence of learning rates, and $\mt$ represents the memory vector that
accumulates the elements of the stochastic gradient that have been suppressed by the operator $\phik(\cdot)$.

    \section{Ergodic convergence}
    \label{sec:convergence}
    In this work, we focus on the memory-less variant of SGD, and we study its convergence properties.
The convergence of sparsified SGD with memory has been studied for both strongly convex \cite{NIPS2018_7697,NIPS2018_7837}
and non-convex objectives \cite{NIPS2018_7837}, but only in the synchronous case.
Nevertheless, in our empirical study, we compare both variants, and verify that
the one with memory considerably benefits from error correction, as expected \cite{NIPS2018_7697}.
Before proceeding with the statement of the main theorem, we formalize our assumptions.
\begin{assumption}
    \label{ass1}
    $\fx$ is continuously differentiable and bounded below:
    \[ \infop\limits_x \fx > - \infty. \]
\end{assumption}
\begin{assumption}
    \label{ass2}
    $\nabla \fx$ is $L$-Lipschitz smooth:
    \[
        \forall \xvect,\yvect \in \Rd \text{,} \left \| \nabla \fx - \nabla \fy \right \| \leq L \left \| \xvect-\yvect \right \|.
    \]
\end{assumption}
\begin{assumption}
    \label{ass3}
    The variance of the (mini-batch) stochastic gradients is bounded:
    \[
        \mathbb{E}\left [ \left \| \mbstgrad - \gradf  \right \|^{2} \right ] \leq \sigma ^{2},
    \]
    where $\sigma ^{2}>0$ is a constant.
\end{assumption}
\begin{assumption}
    \label{ass4}
    Distributed workers might use stale models to compute gradients $\ambstgrad$.
    We assume bounded staleness, that is:
    $t-S\leq\taut \leq t$. In other words, the model staleness $\taut$ satisfies the inequality $t - \taut \leq S$.
    We call $S \geq 0$ the maximum delay.
\end{assumption}
\begin{assumption}
    \label{ass5}
    Let the  \textbf{expected cosine distance} be:
    \begin{equation*}
        \frac{\mathbb{E} \left[ \left< \phik\left( \ambstgrad \right), \gradf \right> \right]}{\mathbb{E} \left[  {\|\phik\left( \ambstgrad \right)\| \, \|\gradf\|}\right]}=\mu_t\geq \mu.
    \end{equation*}
    We assume, similarly to previous work \cite{Dai:2019}, that the constant $\mu > 0$ measures the discrepancy between the sparsified stochastic gradient and the full gradient.
\end{assumption}

\begin{theorem}
    \label{theor:convergence}
    Let Assumptions \ref{ass1}--\ref{ass5} hold.
    Consider the memory-less sparsified SGD defined in \Cref{algo:memory-less}, which uses the
    $\phik(\cdot)$ top-$k$ operator for a given $1 \leq k \leq d$.
    Then, for an appropriately defined learning rate $\etat = \frac{\rho\mu}{L \sqrt{t+1}}$ and for $\Lambda = \left( f \left ( \xvect_{0} \right ) - \infop\limits_{\xvect} \fx  \right)$,
    it holds that:
    \begin{flalign*}
        &\minop\limits_{0\leq t\leq T}\egrad\leq \frac{\left(\sumT\left(\frac{\etat^2 L}{2}\sigmasq\right)\right)+\Lambda+C}{\sumT\left(\etat\rho\mu-\frac{\etat^2 L}{2}\right)}.
    \end{flalign*}
\end{theorem}
\begin{corollary}\label{corollary:convergence}
Let the conditions of \Cref{theor:convergence} hold.
Then for an appropriately defined constant learning rate $\etat=\eta=\frac{\rho\mu}{L\sqrt{T}}$, we have that:
\begin{flalign*}
&\minop\limits_{0\leq t\leq T}\egrad\leq\left(\frac{\sigmasq}{2}+\frac{(\Lambda+C)L}{(\rho\mu)^2}\right)\frac{1}{\sqrt{T}-\frac{1}{2}}.
\end{flalign*}
\end{corollary}

Asymptotically, the convergence rate of memory-less sparsified SGD behaves as
$\mathcal{O}(\frac{1}{\sqrt{T}})$, which matches the best known results for non-convex
SGD \cite{ghadimi2013stochastic}, and for non-convex
asynchronous SGD \cite{Lian:2015:APS:2969442.2969545}.
Note that considering a constant learning rate intuitively makes sense. 
When gradients are not heavily sparsified, i.e., $\rho$ is large, we can afford a large learning rate. 
Similarly, when stale, sparse stochastic, and full gradients are similar, i.e., when $\mu$ is large, we can again set a large learning rate.

It is more difficult to quantify the role of the constant terms in \Cref{corollary:convergence}, especially
those involving sparsification.
While it is evident that aggressive sparsification (extremely small $\rho$) could harm convergence, the exact role
of the second constant term heavily depends on the initialization and the geometry of the loss function, which we
do not address in this work.
We thus resort to a numerical study to clarify these questions, but introduce a proxy for measuring convergence rate.
Instead of imposing a target test accuracy, and use the number of training iterations to measure convergence rate
(which we found to be extremely noisy), we fix an iteration budget, and measure the test accuracy once training concludes.\\

\noindent \textbf{Remarks.} A careful assessment of \Cref{ass5} is in order.
We assume that a sparse version of a stochastic gradient computed with respect to a stale model,
does not diverge too much from the true, full gradient%
\footnote{A similar remark, albeit without sparsification, has been made in \cite{Dai:2019}.}.
We measure this coherency trough a positive constant $\mu > 0$.
However, it is plausible to question the validity of such assumption, especially in a situation where either the
sparsification is too aggressive, or the maximum delay is too high.

\begin{figure}
    \hspace{-16pt}
    \subfloat[
        Expected cosine similarity $\mu_t$
    ]{
        \small
        \pgfplotsset{height=4.0cm}
        \pgfplotsset{every x tick label/.append style={font=\fontsize{4}{4}\selectfont}}
        \pgfplotsset{every y tick label/.append style={font=\fontsize{6}{4}\selectfont}}
        \input{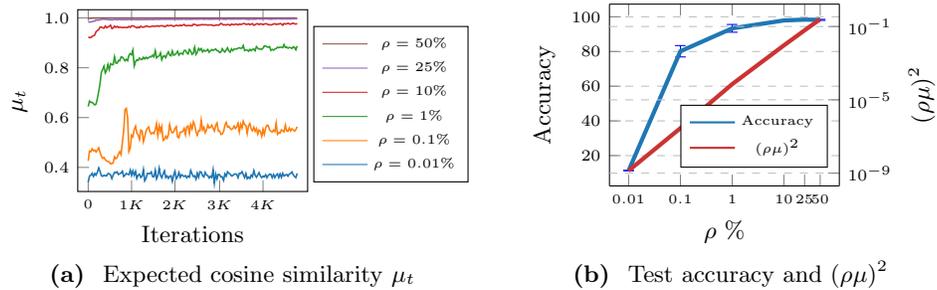}
        \label{fig:remarks_a}
    }
    \subfloat[
        Test accuracy and $(\rho\mu)^2$
    ]{
        \pgfplotsset{height=4.0cm}
        \pgfplotsset{every x tick label/.append style={font=\fontsize{4}{4}\selectfont}}
        \pgfplotsset{every y tick label/.append style={font=\fontsize{6}{4}\selectfont}}
        \begin{tikzpicture}
\definecolor{color0}{rgb}{0.12156862745098,0.466666666666667,0.705882352941177} 
\definecolor{color3}{rgb}{0.8,0.2,0.2} 
     \begin{semilogxaxis}[
     axis y line*=left,
       xlabel={$\rho$ \%},
    ylabel={Accuracy},
    xticklabels={{$0.01$}, {$0.1$},{$1$},{$10$},{$25$},{$50$}},
    xtick = {0.01,0.1,1,10,25,50},
     ymajorgrids=true,
    grid style=dashed,
    every axis plot/.append style={thick}
     ]
     \addlegendimage{no markers, color0}
     
       \addplot+[mark=none, error bars/.cd, y dir=both,y explicit] coordinates{
        (0.01,11.364) +-(0,0.118)   (0.1,80.138)  +-(0,3.271) (1,93.352) +-(0,2.233)(10,97.902) +-(0,0.317) (25,98.364) +-(0,0.194) (50,98.248) +-(0,0.302)
        };

        \addplot[mark=none, ultra thick, color0] coordinates{
        (0.01,11.364) +-(0,0.118)   (0.1,80.138)  +-(0,3.271) (1,93.352) +-(0,2.233)(10,97.902) +-(0,0.317) (25,98.364) +-(0,0.194) (50,98.248) +-(0,0.302)
        };
     \end{semilogxaxis}
     \begin{loglogaxis}[
      hide x axis,
       axis y line*=right,
       xlabel={$\rho$ \%},
    ylabel={$(\rho\mu)^2$},
    xticklabels={{$0.01$}, {$0.1$},{$1$},{$10$},{$25$},{$50$}},
    xtick = {0.01,0.1,1,10,25,50},
     ymajorgrids=true,
    grid style=dashed,
    every axis plot/.append style={thick},
    legend style={font=\tiny},
    legend pos=south east,
    legend style={at={(0.95,0.1)}}
     ]
     \addlegendimage{no markers, color0}
     \addlegendimage{no markers, color3}
      \addplot[mark=none, ultra thick, color3] coordinates{
        (0.01,0.0000000013619)  (0.1,0.00000028447)  (1,0.000071684) (10,0.0094) (25,0.0618) (50,0.2499)};
        
    \addlegendentry{Accuracy}
    \addlegendentry{$(\rho\mu)^2$}
     \end{loglogaxis}
   \end{tikzpicture}
        \label{fig:remarks_b}
    }
    \caption{Empirical results in support to \Cref{ass5}. Experiments for \ssgd with \lenet on \mnist,
    using a range of possible sparsification coefficients $\rho$.}
    \label{fig:remarks}
\end{figure}

%
%

We study the limits of our assumption empirically, and report our findings in \Cref{fig:remarks}.
The evolution of the expected cosine similarity $\mu_t$ defined in \Cref{ass5}, reported here as a function of
algorithmic progress, is in line with our assumption.
Clearly, aggressive sparsification negatively impacts gradient coherency, as shown in \Cref{fig:remarks_a}.
Moreover, as expected from \Cref{theor:convergence}, convergence rate measured through the proxy of test accuracy,
also increases with $(\rho\mu)^2$.
When sparsification is too aggressive, $(\rho\mu)^2$ is too small, which harms convergence.

\subsection{Proof Sketch}
\label{subsec:sketch}
We now give an outline of the proof of Theorem~\ref{theor:convergence},
whereas the full proof is available in \Cref{sec:th_proof}.
Following standard practice in non-convex asynchronous settings \cite{Liu:2015,Lian:2015:APS:2969442.2969545},
we settle for the weaker notion of ergodic convergence to a local minimum of the function~$f$.
Our strategy is to bound the expected sum-of-squares gradients of~$f$.
By the $L$-Lipshitz property of $\nabla f \left ( x \right )$ (see \Cref{ass2}), we have that:
\begin{flalign}
    \label{eq:begin}
    f\left ( \xvect_{t+1} \right ) & \leq \fxt + \left \langle \xvect_{t+1} - \xt, \gradf \right \rangle + \frac{L}{2} \left \| \xvect_{t+1} - \xt \right \|^{2} \nonumber \\
    & = \fxt - \etat \left \langle \phik \left [ \ambstgrad \right ], \gradf \right \rangle + \frac{\etat^{2} L}{2} \left \| \compresgrad \right \|^{2}.
\end{flalign}
The strategy to continue the proof is to find an upper bound for the term $\mathbb{E} \left[\left \| \compresgrad \right \|^{2}\right]$
and a lower bound for the term $\mathbb{E} \left[\left \langle \phik \left [ \ambstgrad \right ], \gradf \right \rangle\right]$.

Let's focus on the term $\etat \left \langle \phik \left [ \ambstgrad \right ], \gradf \right \rangle$.
Using \Cref{lemma:comp2vect}, \Cref{ass5}, and some algebraic manipulations, we can bound the expectation of
the above term as follows:
\[
     \mathbb{E} \left[ \etat \left \langle \compresgrad , \gradf \right \rangle  \right] \geq \etat \rho \mu \egrad,
\]
where $\rho = k / d$, and $\mu$ is defined in \Cref{ass5}.

Next, we can bound the expectation of the term $\frac{\etat^{2} L}{2} \left \| \compresgrad \right \|^{2}$ by
remarking that:
\[\mathbb{E} \left[ \left \| \compresgrad \right \|^{2} \right] \leq \mathbb{E} \left[ \left \| \gradftau \right \|^{2} \right ] + \sigmasq.\]
We then introduce a bound for the term:
\[  \sumT \eta^2_t \mathbb{E} \left[ \left \| \gradftau \right \|^{2} \right ]\leq\sumT \eta^2_t \egrad+C,\]
where $C$ is a positive finite constant.

Finally, if we take the expectation of the whole inequality \ref{eq:begin}, sum over $t$ from 0 to $T-1$, use
\Cref{ass1} and \Cref{ass5}, the derivations above, and let $\Lambda = \left( f \left ( \xvect_{0} \right ) - \infop\limits_{\xvect} \fx  \right)$, by rearranging we obtain:
\begin{equation*}
    \sumT \left ( \etat\rho\mu - \frac{L \etat^{2}}{2} \right ) \egrad \leq \Lambda  + C+ \frac{\sigmasq L}{2}\sumT\etat^{2}.
\end{equation*}
from which we derive the result of \Cref{theor:convergence}:
\[
\minop\limits_{0\leq t\leq T}\egrad\leq \frac{\left(\sumT\left(\frac{\etat^2 L}{2}\sigmasq\right)\right)+\Lambda+C}{\sumT\left(\etat\rho\mu-\frac{\etat^2 L}{2}\right)}.
\]
Moreover, by choosing an appropriate constant learning rate ($\etat = \frac{\rho \mu}{L \sqrt{T}}$), we can derive
\Cref{corollary:convergence}:
\begin{flalign*}
&\minop\limits_{0\leq t\leq T}\egrad\leq\left(\frac{\sigmasq}{2}+\frac{(\Lambda+C)L}{(\rho\mu)^2}\right)\frac{1}{\sqrt{T}-\frac{1}{2}}.
\end{flalign*}


    \section{Experiments}
    \label{sec:experiments}
While the benefits of sparsification have been extensively validated in the literature \cite{DBLP:journals/corr/AjiH17,7835789,seide20141,strom2015scalable,NIPS2017_6768},
such works focus on communication costs in a synchronous setup, with the exception of the work in \cite{NIPS2018_7697}, which illustrates
a simple experiment in a multi-core asynchronous setup.
Instead, our experiments focus on verifying that: 1) the effects of staleness are negligible; 2) sparsification
does not harm convergence rates, using test accuracy 
as a proxy; 3) the benefits
of the memory mechanism applied to sparsified SGD.
We consider several worker delay distributions, and we compare the performance of the three SGD variants:
standard SGD, and sparsified SGD with and without memory.
We also investigate the effects of scaling-out the system, by going up to 128 workers.

For our experimental campaign, we have built a custom simulator that plugs into existing machine learning libraries to
leverage automatic differentiation and the vast availability of models, but abstracts away the complications of a real
distributed setting.
With this setup, it is easy to compare a variety of stochastic optimization algorithms on realistic and complex loss
functions.
More details about our simulator are given in \Cref{sec:sim_staleness}.

\subsection{Experimental setup}

\paragraph{SGD variants.} We compare sparsified SGD without (\ssgd) and with memory (\ssgdm) to ``vanilla'' asynchronous SGD (\vanilla).
For all algorithms, and for all scenarios, we perform a grid search to find the best learning rate.
When relevant, Figures report standard deviation, obtained by repeating our experiments 5 times.
Note that for direct comparisons on individual experiments to be fair, we make sure to use the same algorithmic
initialization for SGD (e.g., we use the same initial model parameters), and the same simulation seed.


\paragraph{Parameters.}
We configure the system architecture as follows: we consider a ``parameter server'' setup, whereby a given number of worker machines
are connected to a master by a simple network model, we do not simulate network congestion, we impose fair bandwidth
sharing, and we do not account for routing overheads.

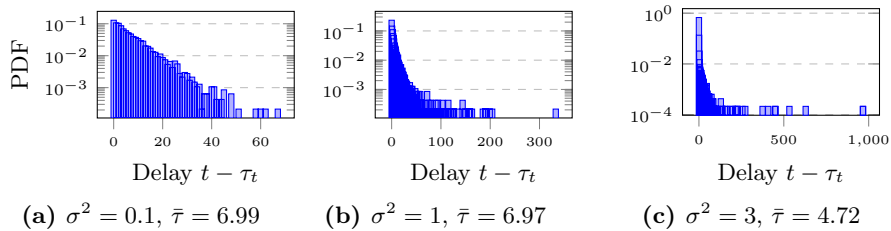
\begin{figure*}[!t]
        \hspace{-1.2em}
        \subfloat[$\sigmasq=0.1$, $\avtau=6.99$]{%
        \small
        \pgfplotsset{width=4.2cm}
        \pgfplotsset{height=3.0cm}
        \pgfplotsset{every x tick label/.append style={font=\fontsize{4}{4}\selectfont}}
        \pgfplotsset{every y tick label/.append style={font=\fontsize{6}{4}\selectfont}}
        \begin{tikzpicture}
    \begin{axis}[
    ymode=log,
    log origin=infty,
    xlabel={Delay $t-\tau_t$},
    ylabel={PDF},
    y tick label style={
    /pgf/number format/fixed,
    /pgf/number format/precision=1,
    },
    ybar,
    bar width=2pt,
    xtick pos=left,
    ymajorgrids=true,
    grid style=dashed,
    ]

\addplot plot coordinates
            {(0,0.128785) (1,0.106610) (2,0.103412) (3,0.086780) (4,0.069083) (5,0.060554) (6,0.052452) (7,0.047761) (8,0.040085) (9,0.037527) (10,0.034328) (11,0.029424) (12,0.028571) (13,0.020256) (14,0.019616) (15,0.017271) (16,0.013433) (17,0.012793) (18,0.009382) (19,0.011301) (20,0.008742) (21,0.007463) (22,0.005330) (23,0.006610) (24,0.004264) (25,0.006823) (26,0.005544) (27,0.002772) (28,0.002132) (29,0.002985) (30,0.002772) (31,0.001706) (32,0.002132) (33,0.001493) (34,0.001066) (35,0.001279) (36,0.000213) (37,0.000640) (38,0.000640) (40,0.001066) (41,0.001066) (42,0.000426) (43,0.000426) (44,0.000426) (45,0.000853) (48,0.000640) (51,0.000213) (58,0.000213) (60,0.000213) (62,0.000213) (67,0.000213)};

    \end{axis}
\end{tikzpicture}\label{fig:var01}
        }
         \hspace{-1.2em}
        \subfloat[$\sigmasq=1$, $\avtau=6.97$]{%
        \small
        \pgfplotsset{width=4.2cm}
        \pgfplotsset{height=3.0cm}
        \pgfplotsset{every x tick label/.append style={font=\fontsize{4}{4}\selectfont}}
        \pgfplotsset{every y tick label/.append style={font=\fontsize{6}{4}\selectfont}}
        \begin{tikzpicture}
    \begin{axis}[
    ymode=log,
    log origin=infty,
    xlabel={Delay $t-\tau_t$},
    y tick label style={
    /pgf/number format/fixed,
    /pgf/number format/precision=1,
    },
    ybar,
    bar width=2pt,
    xtick pos=left,
    ymajorgrids=true,
    grid style=dashed,
    ]

\addplot plot coordinates
            {(0,0.232196) (1,0.145629) (2,0.110235) (3,0.081663) (4,0.070576) (5,0.053305) (6,0.038806) (7,0.032409) (8,0.028358) (9,0.022388) (10,0.018550) (11,0.016844) (12,0.014072) (13,0.014925) (14,0.010021) (15,0.007036) (16,0.008529) (17,0.006610) (18,0.005970) (19,0.007036) (20,0.004904) (21,0.003412) (22,0.005330) (23,0.004051) (24,0.003838) (25,0.002559) (26,0.003198) (27,0.002985) (28,0.002132) (29,0.002132) (30,0.001279) (31,0.001919) (32,0.000853) (33,0.001279) (34,0.000640) (35,0.001279) (36,0.001279) (37,0.001706) (38,0.001066) (40,0.001279) (41,0.000640) (42,0.001279) (43,0.000640) (44,0.000640) (45,0.000640) (48,0.001066)
            (49,0.000640) (50,0.000853) (51,0.000213) (52,0.000853)
            (53,0.000426) (54,0.000213) (55,0.000853) (56,0.000213) (57,0.000213) (58,0.000853) (60,0.000426) (61,0.000426)
            (62,0.000213) (64,0.000213) (65,0.000213) (66,0.000213) (67,0.000213) (68,0.000426) (69,0.000213) (70,0.000213)
            (72,0.000853) (73,0.000213) (74,0.000213) (75,0.000426) (76,0.000213) (79,0.000213) (80,0.000426) (81,0.000213)
            (84,0.000213) (85,0.000213) (86,0.000213) (87,0.000213) (88,0.000213) (89,0.000426) (91,0.000213) (92,0.000213)
            (93,0.000213) (94,0.000426) (97,0.000213) (100,0.000213)
            (103,0.000426) (104,0.000213) (91,0.000213) (105,0.000213)
            (107,0.000213) (110,0.000213) (114,0.000213) (115,0.000426)
            (116,0.000213) (119,0.000213) (121,0.000213) (124,0.000213)
            (125,0.000213) (126,0.000213) (137,0.000213) (144,0.000426)
            (149,0.000213) (150,0.000213) (157,0.000213) (158,0.000213)
            (159,0.000213) (163,0.000213) (185,0.000213) (190,0.000213)
            (191,0.000213) (199,0.000213) (204,0.000213) (333,0.000213)};

    \end{axis}
\end{tikzpicture}\label{fig:var1}
        }
        \subfloat[$\sigmasq=3$, $\avtau=4.72$]{%
        \small
        \pgfplotsset{width=4.2cm}
        \pgfplotsset{height=3.0cm}
        \pgfplotsset{every x tick label/.append style={font=\fontsize{4}{4}\selectfont}}
        \pgfplotsset{every y tick label/.append style={font=\fontsize{6}{4}\selectfont}}
        \begin{tikzpicture}
    \begin{axis}[
    ymode=log,
    log origin=infty,
    xlabel={Delay $t-\tau_t$},
    y tick label style={
    /pgf/number format/fixed,
    /pgf/number format/precision=1,
    },
    ybar,
    bar width=2pt,
    xtick pos=left,
    ymajorgrids=true,
    grid style=dashed,
    ]

\addplot plot coordinates
            {(0,0.673561) (1,0.135181) (2,0.032196) (3,0.01472) (4,0.007676) (5,0.008316) (6,0.009595) (7,0.006397) (8,0.007463) (9,0.005330) (10,0.004904) (11,0.005544) (12,0.004478) (13,0.004264) (14,0.004051) (15,0.005544) (16,0.003198) (17,0.003412) (18,0.003412) (19,0.003838) (20,0.001706) (21,0.003412) (22,0.002985) (25,0.001066) (24,0.002559) (25,0.001066) (26,0.002559) (27,0.002345) (28,0.001279) (29,0.001706) (30,0.000853) (31,0.001066) (32,0.001279) (33,0.000213) (34,0.001493) (35,0.001279) (36,0.001066) (37,0.001066) (38,0.000426) (39,0.000853) (40,0.001066) (41,0.001066) (42,0.000853) (43,0.001493) (44,0.000853) (45,0.001066) (46,0.001706) (47,0.000640) (48,0.000640) (49,0.000640) (51,0.000853) (52,0.000426) (53,0.000640) (54,0.000641) (55,0.000426) (56,0.000640) (57,0.000426) (58,0.000213) (59,0.000213) (60,0.000640) (61,0.000640) (62,0.000213) (64,0.000213) (65,0.000213) (66,0.000426) (67,0.000213) (68,0.000426) (70,0.000213) (72,0.000213) (74,0.000640) (76,0.000426) (77,0.000213) (78,0.000426) (80,0.000213) (81,0.000213) (83,0.000213) (85,0.000426) (87,0.000213) (96,0.000213) (98,0.000213) (113,0.000213) (129,0.000426)
            (136,0.000213) (138,0.000213) (154,0.000213) (164,0.000213)
            (185,0.000213) (200,0.000213) (216,0.000213) (223,0.000213)
            (229,0.000213) (256,0.000213) (273,0.000213) (375,0.000213)
            (408,0.000213) (448,0.000213) (451,0.000213) (533,0.000213)
            (629,0.000213) (966,0.000213) (967,0.000213)};

    \end{axis}
\end{tikzpicture}\label{fig:var3}
        }
        \caption{Delay distributions of a simulation run with \lenet on \mnist, in a distributed setting with 8 workers. For each worker we generate a network delay according to an exponential distribution with rate $\lambda$. We sample $\lambda$ from a log-normal distribution with mean 0 and variance $\sigmasq$. For each configuration, we also report the resulting average staleness $\avtau$.
        }
        \label{fig:staleness_distributions}
    \end{figure*}

In our simulations, both computation and communication costs can be modeled according to a variety of
distributions.
In this work we use uniformly distributed computation times with a small support, that are indicative of an homogeneous system.
Instead of directly controlling the staleness of gradient updates, as done in other studies
\cite{Damaskinos:2018,Dai:2019}, we indirectly induce staleness by imposing synthetic network delays,
which we generate according to an exponential distribution with rate $\lambda$ (the inverse of the mean).
In particular, each worker samples a value for $\lambda$ from a log-normal distribution with mean 0 and variance $\sigmasq$.
\Cref{fig:staleness_distributions} shows the resulting delay distribution for the entire training period in a simulation with 8 workers,
using different values of $\sigmasq$.
As we increase $\sigmasq$, the maximum delay experienced by the workers increases, up to very large values. In addition, the mass of the distribution
shifts towards lower delays; indeed, for higher values of $\sigmasq$, the majority of workers have small delays and only few workers experience very large delays.
This is confirmed by the average staleness $\avtau$, which decreases as $\sigmasq$ increases.
Notice that the interplay between communication delay and staleness is subtle: we provide a comprehensive
description of the staleness generation process in \Cref{sec:sim_staleness}, with
illustrations that help understanding the shape of the $\taut$ distribution.

\paragraph{Models and datasets.} We consider a classification task, where we train two
convolutional neural network (\cnn) variants of 
increasing model complexity, to gain insight on the role of sparsification for large deep network models.
First, we study the behavior of \lenet, using the \mnist dataset, then we move on to \resnet, using the \cifar dataset.
The model parameter and gradients dimensionality are approximately $d \in \{60K, 600K\}$ for
\lenet and \resnet, respectively.
We use a training mini-batch size of 64 for \mnist and 128 for \resnet and a testing mini-batch size of 128 samples.
Additional details are available in \Cref{sec:experiment_details}.

\subsection{Comparative analysis}\label{subsec:comparative_analysis}
We compare \ssgd and \ssgdm with \vanilla by measuring the test accuracy reached after a fixed number of epochs,
which we set to 5 for \lenet on \mnist and 161 for \resnet on \cifar.
We consider three scenarios, with 8 workers: each has a different delay distribution, given by the parameter $\sigmasq$,
as shown in \Cref{fig:staleness_distributions}.
For sparsified methods, we use the best sparsification coefficient $\rho$. We discuss how $\rho$ can be tuned in \Cref{subsec:details}.



\begin{figure*}[!ht]
  
  \begin{center}
        \centering
        \small
        \pgfplotsset{width=6.0cm}
        \pgfplotsset{height=4.0cm}
        \pgfplotsset{every x tick label/.append style={font=\fontsize{4}{4}\selectfont}}
        \pgfplotsset{every y tick label/.append style={font=\fontsize{6}{4}\selectfont}}
        \input{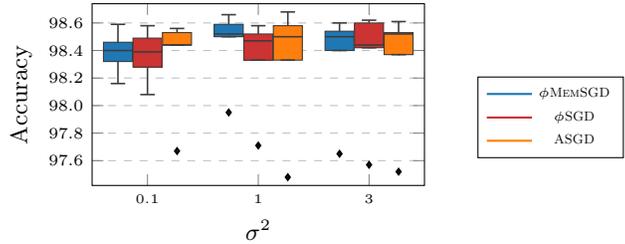}
    \end{center}

    \caption{Comparison of test accuracy of \lenet on \mnist,
    for three different asynchronous settings with 8 workers.
    In each setting we sample the exponential rates $\lambda$ from a log-normal distribution with mean 0 and variance $\sigmasq$. For sparsified methods, the best $\rho$ has been taken.}
    \label{fig:comparison_all}
\end{figure*}

\begin{table}[!t]
\centering
    \begin{tabular}{|c|c|c|}
        \ssgdm & \ssgd   & \vanilla\\
        \hline
        87.05 $\pm$ 0.53 & 86.21 $\pm$ 1.06 & 85.90 $\pm$ 1.01 \\
    \end{tabular}
    \caption{Comparison of test accuracy of \resnet on \cifar, with $\sigmasq=0.1.$}
    \label{tab:resnet_results}
\end{table}

\Cref{fig:comparison_all} illustrates results obtained using a \lenet architecture with the \mnist dataset, while \Cref{tab:resnet_results} reports the results obtained with \resnet on \cifar, fixing $\sigmasq=0.1$.

Clearly, for both simple and deep models, the effects of sparsification on test accuracy are negligible.
Given a reasonable choice of sparsification $\rho$, all variants achieve similar test accuracy using the same number
of training iterations.
Moreover, sparsified methods consistently achieve comparable performance to the non-sparse method irrespectively of the delay
distribution.
This confirms the result in \Cref{corollary:convergence}, which indicates asymptotically vanishing
effects of staleness on convergence (indeed, the term $\tau$ does not appear in the bound).
We also observe that, as expected, the memory-based variant of sparsified SGD has an edge on the memory-less method,
because it achieves better performance for lower values of $\rho$.
This results clarifies the impact of memory-based error correction as a method
to recover lost information due to aggressive sparsification.

Finally, note that we explicitly do not compare the methods using
wall-clock times, as we are not interested in measuring the well-known benefits of sparsification in terms of reduced communication costs.

\subsection{Tuning gradient sparsification}\label{subsec:details}

Using \cref{theor:convergence} alone, it can be difficult to understand how $\rho$ can be tuned.
Next, we focus on the \lenet architecture using the \mnist, trained for 5 epochs, to understand how this affects the accuracy of sparsified SGD.

\begin{figure}[!ht]
    \centering
    \pgfplotsset{width=6.0cm}
    \pgfplotsset{height=4.0cm}
    \pgfplotsset{every x tick label/.append style={font=\fontsize{4}{4}\selectfont}}
    \pgfplotsset{every y tick label/.append style={font=\fontsize{6}{4}\selectfont}}
    \begin{tikzpicture}
    \definecolor{color0}{rgb}{0.12156862745098,0.466666666666667,0.705882352941177} 
    \definecolor{color3}{rgb}{0.8,0.2,0.2} 

    \begin{semilogxaxis}[
    xlabel={$\rho$ \%},
    ylabel={Accuracy},
    xticklabels={{$0.01$}, {$0.1$},{$1$},{$10$},{$25$},{$50$}},
    xtick = {0.01,0.1,1,10,25,50},
    ymajorgrids=true,
    grid style=dashed,
    every axis plot/.append style={thick},
    legend style={font=\tiny},
    legend pos=south east,
    legend style={at={(1.5,0.3)}},
    ]

        \addlegendimage{no markers, color0}
        \addlegendimage{no markers, color3}

        \addplot+[mark=nonne, error bars/.cd, y dir=both,y explicit] coordinates {
        (0.01,97.776) +-(0,0.127)   (0.1,98.166) +-(0,0.136)  (1,98.344) +-(0,0.197) (10,98.380) +-(0,0.194) (25,98.330) +-(0,0.150) (50,98.386) +-(0,0.160)
        };

        \addplot[mark=none, ultra thick, color0, forget plot] coordinates {
        (0.01,97.776) +-(0,0.127)   (0.1,98.166) +-(0,0.136)  (1,98.344) +-(0,0.197) (10,98.380) +-(0,0.194) (25,98.330) +-(0,0.150) (50,98.386) +-(0,0.160)
        };

        \addplot+[mark=none, error bars/.cd, y dir=both,y explicit] coordinates{
        (0.01,11.364) +-(0,0.118)   (0.1,80.138)  +-(0,3.271) (1,93.352) +-(0,2.233)(10,97.902) +-(0,0.317) (25,98.364) +-(0,0.194) (50,98.248) +-(0,0.302)
        };

        \addplot[mark=none, ultra thick, color3, forget plot] coordinates{
        (0.01,11.364) +-(0,0.118)   (0.1,80.138)  +-(0,3.271) (1,93.352) +-(0,2.233)(10,97.902) +-(0,0.317) (25,98.364) +-(0,0.194) (50,98.248) +-(0,0.302)
        };

        \legend{\ssgdm, \ssgd}
    \end{semilogxaxis}
\end{tikzpicture}

    \caption{Detailed study to understand how to tune the sparsification $\rho$ , for \lenet on \mnist. Test accuracy as a function of $\rho$, in a system with 8 workers and $\sigmasq=0.1$}
    \label{fig:tuning}
\end{figure}
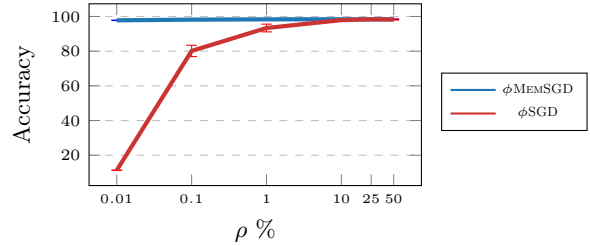

\begin{figure}[!ht]
    \centering
    \pgfplotsset{width=6.0cm}
    \pgfplotsset{height=4.0cm}
    \pgfplotsset{every x tick label/.append style={font=\fontsize{4}{4}\selectfont}}
    \pgfplotsset{every y tick label/.append style={font=\fontsize{6}{4}\selectfont}}
    \begin{tikzpicture}
    \definecolor{color0}{rgb}{0.12156862745098,0.466666666666667,0.705882352941177} 
    \definecolor{color1}{rgb}{1,0.498039215686275,0.0549019607843137} 
    \definecolor{color2}{rgb}{0.372549019607843,0.72156862745098,0.356862745098039} 
    \definecolor{color3}{rgb}{0.8,0.2,0.2} 

    \begin{semilogxaxis}[
    xlabel={Workers},
    ylabel={Accuracy},
    xticklabels={{$1$}, {$2$},{$4$},{$8$},{$16$},{$32$},{$64$},{$128$}},
    xtick = {1,2,4,8,16,32,64, 128},
    legend pos=south east,
    ymajorgrids=true,
    grid style=dashed,
    every axis plot/.append style={thick},
    legend style={at={(1.6,0.05)}},
    legend style={font=\tiny}
    ]

        \addlegendimage{no markers, color0}
        \addlegendimage{no markers, color3}
        \addlegendimage{no markers, color1}

        \addplot+[mark=none, error bars/.cd, y dir=both, y explicit] coordinates {
        (1,98.344) +- (0,0.206)   (2,98.466) +- (0,0.151)   (4,98.480) +- (0,0.137)  (8,98.444) +- (0,0.283)   (16,98.438) +- (0,0.205)    (32,98.296) +- (0,0.222)    (64,98.258) +- (0,0.198)   (128,98.012) +- (0,0.247)
        };
    \addlegendentry{\ssgdm}
        \addplot[mark=none, ultra thick, color0, forget plot] coordinates {
        (1,98.344) +- (0,0.206)   (2,98.466) +- (0,0.151)   (4,98.480) +- (0,0.137)  (8,98.444) +- (0,0.283)   (16,98.438) +- (0,0.205)    (32,98.296) +- (0,0.222)    (64,98.258) +- (0,0.198)   (128,98.012) +- (0,0.247)
        };
        \addplot+[mark=none, error bars/.cd, y dir=both,y explicit] coordinates {
        (1,98.302) +- (0,0.157)   (2,98.314) +- (0,0.331)   (4,98.420) +- (0,0.142)  (8,98.322) +- (0,0.354)   (16,98.338) +- (0,0.153)    (32,98.262) +- (0,0.234)    (64,97.996) +- (0,0.090)  (128,97.978) +- (0,0.257)
        };
        \addlegendentry{\ssgd}
        \addplot[mark=none, ultra thick, color3, forget plot] coordinates {
        (1,98.302) +- (0,0.157)   (2,98.314) +- (0,0.331)   (4,98.420) +- (0,0.142)  (8,98.322) +- (0,0.354)   (16,98.338) +- (0,0.153)    (32,98.262) +- (0,0.234)    (64,97.996) +- (0,0.090)   (128,97.978) +- (0,0.257)
        };

        \addplot+[mark=none, error bars/.cd, y dir=both,y explicit] coordinates {
        (1,98.304) +- (0,0.185)   (2,98.406) +- (0,0.281)   (4,98.308) +- (0,0.258)  (8,98.314) +- (0,0.483)   (16,98.280) +- (0,0.287)    (32,98.182) +- (0,0.294)    (64,97.720) +- (0,0.601)  (128,97.742) +- (0,0.540)
        };
       \addlegendentry{\vanilla}
        \addplot[mark=none, ultra thick, color1, forget plot] coordinates {
        (1,98.304) +- (0,0.185)   (2,98.406) +- (0,0.281)   (4,98.308) +- (0,0.258)  (8,98.314) +- (0,0.483)   (16,98.280) +- (0,0.287)    (32,98.182) +- (0,0.294)    (64,97.720) +- (0,0.601)   (128,97.742) +- (0,0.540)
        };



        \legend{\ssgdm,\ssgd, \vanilla}
    \end{semilogxaxis}
\end{tikzpicture}

    \caption{Comparison of test accuracy, as a function of the number of workers. Results for \lenet on \mnist.}
    \label{fig:comparison_workers}
\end{figure}
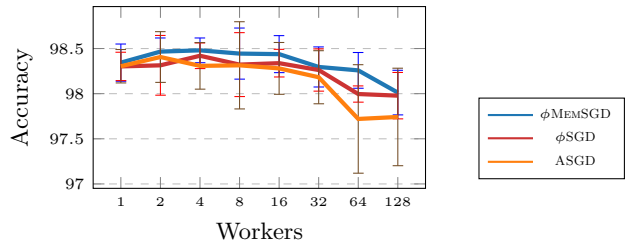

The results in \Cref{fig:tuning} show the impact of different values of the sparsification coefficient $\rho$ on test accuracy.
We notice a stark difference between \ssgdm and \ssgd: the latter is much more sensitive to appropriate choices of $\rho$,
and requires larger coefficients to achieve a reasonable test accuracy.
For \ssgdm, instead, aggressive sparsification doesn't penalize performance noticeably, thanks to the memory mechanism.
Then, even aggressive sparsification can be viable, as the cost per iteration (in terms of transmission times)
decreases drastically, compared to standard SGD.
Also, note that the top-$k$ operator can be executed efficiently on GPUs \cite{shanbhag2018efficient},
so that computational costs per iteration are equivalent to standard SGD.


\subsection{Scalability}\label{subsec:scalability}

We now investigate how the three SGD variants scale with an increasing number of workers.
As shown in \Cref{fig:comparison_workers}, all variants of SGD incur a slight drop in performance
with more workers.
Indeed, with more workers, the delay distribution in the system changes according to \Cref{fig:staleness_distributions},
due to an increase in the probability of picking large delays.
We note again that applying sparsification does not harm the performance of SGD, in that both \ssgdm and \ssgd reach a comparable
test accuracy with respect to \vanilla.
This is valid also for a deep convolutional network: for example, we run \resnet on \cifar with 32 workers and we obtained a test accuracy
of $85.30 \pm 1.24$ for \vanilla and $86.01 \pm 0.49$ for \ssgdm with a sparsification coefficient $\rho=1\%$.

Our results reinforce the message that sparsified SGD should be preferred over vanilla SGD, and this carries over to large scale scenarios
in which, otherwise, excessively large message sizes could entail network congestion and jeopardize algorithmic efficiency.

    \section{Conclusion}
    \label{sec:conclusion}
    In this work we focused on the role of sparsification methods applied to distributed stochastic optimization
of non-convex loss functions, typically found in many modern machine learning settings.

For the first time, we provided a simple and concise analysis of the joint effects of asynchronicity and sparsification
on mini-batch SGD, and showed that it converges asymptotically as $\mathcal{O}\left( 1 / \sqrt{T} \right)$.
Intuitively, top-$k$ sparsification restricts the path taken by model iterates in the optimization landscape to
follow the principal components of a stochastic gradient update, as also noticed in \cite{NIPS2018_7837}.

We complemented our theoretical results with a thorough empirical campaign.
Our experiments covered both variants of sparsified SGD, with and without memory, and compared them to standard SGD.
We used a simple system simulator, which allowed us to explore scenarios with different delay distributions, as wel as
an increasing number of workers.

Our results substantiated the theoretical findings of this work: the effects of staleness vanish asymptotically, and
the impact of sparsification is negligible on convergence rate and test accuracy.
We also studied how to appropriately chose the sparsification factor, and concluded that the memory mechanism
applied to sparsified SGD allows to substantially sparsify gradients, save on communication costs,
while obtaining comparable performance to standard SGD.

Our future plan is to establish a connection between gradient sparsification and recent studies showing that the
landscape of the loss surface of deep models is typically sparse \cite{sagun2017empirical,draxler2018essentially,karakida2018universal}.
In light of such works,  \cite{conf/iclr/FrankleC19} suggests that sparsification can be directly applied to model parameters,
albeit training requires multiple stages.
Gradient sparsification could then be studied as a mechanism to favor model compression at training time.





    \bibliographystyle{splncs04}
    \bibliography{bibliography}
    
       \label{sec:appendix}
    \appendix

\section{Proof of Theorem}\label{sec:th_proof}
    In this section we build all the useful tools to formalize the proof of \Cref{theor:convergence}. In \Cref{sec:comp2vect} we prove the $k$-contraction lemma, in \Cref{assrecap} we restate assumptions for simplicity, we derive useful facts in \Cref{usefulfacts}, derive a tighter bounding term for the sum of magnitudes of stale gradients in \Cref{bounding}, and finally derive the full proof of the convergence theorem in \Cref{proofderiv}.
    \subsection{Proof of \Cref{lemma:comp2vect}}
    \label{sec:comp2vect}
    Given a vector $\uvect \in \mathbb{R}^{d}$, a parameter $1 \leq k \leq d$, and the top-$k$ operator $\phik(\uvect):
    \mathbb{R}^{d} \to \mathbb{R}^{d}$ defined in \Cref{def:topk}, we have that:

    \[
        \|\phik(\uvect)\|^2 \geq \frac{k}{d} \|\uvect\|^2
    \]
    In fact we can write $\|\uvect\|^2$ as follows:
    \begin{equation}
        \begin{split}
            \|\phik(\uvect)\|^2 &= \|\uvect\|^2 - \|\uvect - \phik{\uvect}\|^2 \\
            &\geq \|\uvect\|^2 - \left( 1 - \frac{k}{d} \right) \|\uvect\|^2 \\
            &\geq \frac{k}{d} \|\uvect\|^2
        \end{split}
    \end{equation}
    Where the inequality is obtained by simply applying the k-contraction property.

    \subsection{Recap of Assumptions}\label{assrecap}
    We start by rewriting for simplicity \Cref{ass1} to \Cref{ass5}:
    \begin{enumerate}
        \item
        $\fx$ is continuously differentiable and bounded below: $\infop\limits_x \fx > - \infty$
        \item    $\nabla \fx$ is $L$-Lipschitz smooth:
        \[
            \forall \xvect,\yvect \in \Rd \text{,} \left \| \nabla \fx - \nabla \fy \right \| \leq L \left \| \xvect-\yvect \right \|
        \]
        \item    The variance of the (mini-batch) stochastic gradients is bounded:
        \[
            \mathbb{E}\left [ \left \| \mbstgrad - \gradf  \right \|^{2} \right ] \leq \sigma ^{2},
        \]
        where $\sigma ^{2}>0$ is a constant.

        \item The staleness is bounded, that is:
        $t-S\leq\taut \leq t$, $S\geq 0$. In other words, the model staleness $\taut$ satisfies the inequality $t - \taut \leq S$.
        We call $S \geq 0$ the maximum delay.
        \item The cosine distance between sparse, stale and stochastic gradient and the full one is lower bounded
        \begin{equation*}
            \frac{\mathbb{E} \left[ \left< \phik\left( \ambstgrad \right), \gradf \right> \right]}{\mathbb{E} \left[  {\|\phik\left( \ambstgrad \right)\| \, \|\gradf\|}\right]}=\mu_t\geq \mu
        \end{equation*}
    \end{enumerate}

    Notice moreover that $\xi_{t}$ is statistically independent from $\{\mathbf{x}_0,\cdots,\mathbf{x}_t\}$.

    \subsection{Useful facts}\label{usefulfacts}
    Starting from \Cref{ass2} we can write that:
    \begin{flalign*}
        f(\mathbf{x})\leq f(\mathbf{y})+ \left \langle \mathbf{x}-\mathbf{y}, \nabla f \left ( \mathbf{y} \right ) \right\rangle+ \frac{L}{2} \|\mathbf{x}-\mathbf{y}\|^{2} \quad \forall \mathbf{x},\mathbf{y}
    \end{flalign*}
    Trivially we can rewrite this inequality by using as arguments the two vectors $\xvect_{t},\xvect_{t+1}$:
    \begin{flalign}
        \label{ass2f}
        f\left ( \xvect_{t+1} \right ) & \leq \fxt + \left \langle \xvect_{t+1} - \xt, \gradf \right \rangle + \nonumber \\
        & + \frac{L}{2} \left \| \xvect_{t+1} - \xt \right \|^{2} \nonumber \\
        & = \fxt - \etat \left \langle \phik \left [ \ambstgrad \right ], \gradf \right \rangle \nonumber \\
        & + \frac{\etat^{2} L}{2} \left \| \compresgrad \right \|^{2}
    \end{flalign}
    where the last equality is due to  $\xtp = \xt - \etat \phik\left( \ambstgrad \right)$. Notice that even if $\xvect_t$ as well as $\xi_t$, $\tau_t$ and consequently $f(\xvect_t)$ and $g(\xvect_t)$ are random processes, due to the geometric constraints imposed on the cost function, the above inequality holds with probability 1.

    The second useful quantity we derive is a bound for squared magnitude of $\mbstgrad$. We start with \Cref{ass3}.

    Before proceeding, we introduce the following notation: $\Omega$ is the set of ALL random variables (i.e. $\Omega=\{\xi_0,\cdots,\xi_t,\xvect_0,\cdots,\xvect_t,\tau_0,\cdots,\tau_t\}$), furthermore, we indicate with $\sim \xi_t$ the set difference between $\Omega$ and $\xi_t$.

    We write:
    \begin{flalign*}
        & \mathbb{E}_{\Omega}\left [ \left \| \mbstgrad - \gradf  \right \|^{2} \right ]\\
        &=\mathbb{E}_{\Omega}\left [ \left \| \mbstgrad - \mathbb{E}_{\xi_t}\left(\mbstgrad\right)  \right \|^{2} \right ]\\
        &=\mathbb{E}_{\sim \xi_t}\left [\mathbb{E}_{\xi_t}\left [ \left \| \mbstgrad - \mathbb{E}_{\xi_t}\left(\mbstgrad\right)  \right \|^{2} \right ]\right ]\\
        &=\mathbb{E}_{\sim \xi_t}\left [\mathbb{E}_{\xi_t}\left [ \left \| \mbstgrad \| \right] - \| \mathbb{E}_{\xi_t}\left(\mbstgrad\right)  \right \|^{2} \right ]\\
        &=\mathbb{E}_{\Omega}\left[\| \mbstgrad \| \right]-\mathbb{E}_{\Omega}\left[\| \gradf \| \right]\leq \sigma^2
    \end{flalign*}
    from which:
    \begin{flalign}
        \label{ass3f}
        \mathbb{E}_{\Omega}\left[\| \mbstgrad \| \right]\leq \mathbb{E}_{\Omega}\left[\| \gradf \| \right]+\sigma^2
    \end{flalign}
    \subsection{Bounding magnitudes of delayed gradients}\label{bounding}
    Differently from \cite{Dai:2019}, we derive a tighter bound for the following term:
    \begin{equation}
        \label{newdis}
        \sumT \eta^2_t \mathbb{E} \left[ \left \| \gradftau \right \|^{2} \right ]
    \end{equation}
    Indeed, thanks to the fact that $\etat$ is a decreasing sequence, and using the law of total expectation:
    \begin{flalign*}
        &\sumT \eta^2_t \mathbb{E} \left[ \left \| \gradftau \right \|^{2} \right ]\\&= \sumT \eta^2_t \sum\limits_{l=t-S}^{t}Pr(\tau_t=l)\egradl\\
        &\leq \sumT \sum\limits_{l=t-S}^{t} \eta^2_l Pr(\tau_t=l)\egradl\\
    \end{flalign*}

    Before proceeding, it is useful to introduce a new random quantity, the delay $D$, distributed according to some probability density function $Pr(D=i)=\pi_{i}$. Notice that the true relationship between $Pr(\tau_t)$ and $Pr(D)$ is:
    \begin{flalign*}
        Pr(\tau_t=l)=\frac{\pi_{t-l}}{\sum \limits_{i=0}^{min(t,S)}\pi_i }.
    \end{flalign*}
    Since $t-S\leq\taut \leq t$, obviously the delay variable $D$ has support boundend in $[0,S]$.
    Moreover, to reduce clutter, we define: $\psi_l=\eta^2_l\egradl$.

    Now, we can continue our derivation as:
    \begin{flalign*}
        &\sumT \sum\limits_{l=t-S}^{t}  \frac{\pi_{t-l}}{\sum \limits_{i=0}^{min(t,S)}\pi_i } \psi_l=\\
        &\frac{\pi_{S}\psi_{-S+0}+\pi_{S-1}\psi_{-S+1}+\cdots+\pi_{1}\psi_{\tiny{-1}}+\pi_{0}\psi_{0}
        }{\pi_0}+\\
        &\frac{\pi_{S}\psi_{-S+1}+\pi_{S-1}\psi_{-S+2}+\cdots+\pi_{1}\psi_{0}+\pi_{0}\psi_{1}}{\pi_0+\pi_1}+\\
        &\frac{\pi_{S}\psi_{-S+2}+\pi_{S-1}\psi_{-S+3}+\cdots+\pi_{1}\psi_{1}+\pi_{0}\psi_{2}}{\pi_0+\pi_1+\pi_2}+\\
        &\frac{\pi_{S}\psi_{-S+3}+\pi_{S-1}\psi_{-S+4}+\cdots+\pi_{1}\psi_{2}+\pi_{0}\psi_{3}}{\pi_0+\pi_1+\pi_2+\pi_3}+\\
        &\cdots\\
        &\frac{\pi_{S}\psi_{T-S}+\cdots\cdots\cdots \quad +\pi_{1}\psi_{T-1}+\pi_{0}\psi_{T}}{1}\leq\\
        &\pi_{S}\psi_{-S+0}+\pi_{S-1}\psi_{-S+1}+\cdots+\pi_{1}\psi_{\tiny{-1}}+\pi_{0}\psi_{0}
        +C_0+\\
        &\pi_{S}\psi_{-S+1}+\pi_{S-1}\psi_{-S+2}+\cdots+\pi_{1}\psi_{0}+\pi_{0}\psi_{1}+C_1+\\
        &\pi_{S}\psi_{-S+2}+\pi_{S-1}\psi_{-S+3}+\cdots+\pi_{1}\psi_{1}+\pi_{0}\psi_{2}+C_2+\\
        &\pi_{S}\psi_{-S+3}+\pi_{S-1}\psi_{-S+4}+\cdots+\pi_{1}\psi_{2}+\pi_{0}\psi_{3}+C_3+\\
        &\cdots\\
        &\pi_{S}\psi_{T-S}+\cdots\cdots\cdots \quad +\pi_{1}\psi_{T-1}+\pi_{0}\psi_{T}+0\\
        &\leq \sumT\psi_t+C\\
        &=\sumT \eta^2_t \egrad+C
    \end{flalign*}
    where $C$ is a suitable, finite, constant. We thus proved a strict bound on the sum of magnitudes of delayed gradients as
    \begin{flalign*}
        \sumT \eta^2_t \mathbb{E} \left[ \left \| \gradftau \right \|^{2} \right ]\leq\sumT \eta^2_t \egrad+C.
    \end{flalign*}

    \subsection{Derivation of the theorem}\label{proofderiv}
    We start the derivation from \Cref{ass2f}.
    We rearrange the inequality to bound the increment of cost function at time instant $t$ as:
    \begin{flalign*}
        f\left ( \xvect_{t+1} \right ) -\fxt& \leq - \etat \left \langle \phik \left [ \ambstgrad \right ], \gradf \right \rangle \nonumber \\
        & + \frac{\etat^{2} L}{2} \left \| \compresgrad \right \|^{2}
    \end{flalign*}
    Written in this form, we are still dealing with random quantities. We are interested in taking the expectation of above inequality with respect to \textbf{all} random processes. 
    Then:
    \begin{flalign*}
        & \mathbb{E}_\Omega \left[f\left ( \xvect_{t+1} \right ) -\fxt\right]\\ &\leq  \mathbb{E}_\Omega \left[- \etat \left \langle \phik \left [ \ambstgrad \right ], \gradf \right \rangle
        + \frac{\etat^{2} L}{2} \left \| \compresgrad \right \|^{2}\right]
    \end{flalign*}
    The strategy to continue the proof is to find an upper bound for the expectation term $\mathbb{E}_\Omega \left[\left \| \compresgrad \right \|^{2}\right]$ and a lower bound for the expectation term $\mathbb{E}_\Omega \left[\left \langle \phik \left [ \ambstgrad \right ], \gradf \right \rangle\right]$.
    We start with the upper bound as:
    \begin{flalign*}
        &\mathbb{E}_\Omega \left[\left \| \compresgrad \right \|^{2}\right]\leq \mathbb{E}_\Omega \left[\left \| \ambstgrad \right \|^{2}\right] \\&\leq\mathbb{E} \left[ \left \| \gradftau \right \|^{2} \right ] + \sigmasq,
    \end{flalign*}
    where the first inequality is a trivial consequence of sparsification and the second is the application of \Cref{ass3f}.

    As anticipated, we aim at lower bounding the term: $\mathbb{E}_\Omega \left[\left \langle \phik \left [ \ambstgrad \right ], \gradf \right \rangle\right]$.
    We start with \Cref{ass5} and write:
    \begin{flalign*}
        & \mathbb{E}_{\Omega} \left[ \left< \phik\left( \ambstgrad \right), \gradf \right> \right]\\
        &\geq \mu \mathbb{E}_{\Omega} \left[  {\|\phik\left( \ambstgrad \right)\| \, \|\gradf\|}\right]\\
        &=\mu \mathbb{E}_{\sim \xi_t} \left[ \mathbb{E}_{\xi_t} \left[ {\|\phik\left( \ambstgrad \right)\| \, \|\gradf\|}\right] \right]\\
        &=\mu \mathbb{E}_{\sim \xi_t} \left[ \mathbb{E}_{\xi_t} \left[ \|\phik\left( \ambstgrad \right)\| \right]\, \|\gradf\| \right]\\
    \end{flalign*}
    We focus on the term $\mathbb{E}_{\xi_t} \left[ \|\phik\left( \ambstgrad \right)\| \right]$ and, thanks to the $k$-contraction property and the inequality of the norm of expected values, we can write:
    \begin{flalign*}
        & \mathbb{E}_{\Omega} \left[ \left< \phik\left( \ambstgrad \right), \gradf \right> \right]\\
        &\geq \mu\rho \mathbb{E}_{\sim \xi_t} \left[ \mathbb{E}_{\xi_t} \left[\| \ambstgrad\|\right]\, \|\gradf\| \right]\\
        &\geq \mu\rho \mathbb{E}_{\sim \xi_t} \left[ \|\mathbb{E}_{\xi_t} \left[ \ambstgrad\right]\|\, \|\gradf\| \right]\\
        &=\mu\rho \mathbb{E}_{\sim \xi_t} \left[ \|\gradftau\|\, \|\gradf\| \right]
    \end{flalign*}
    Before proceeding, we reasonably assume that:
    \begin{equation}
        \mathbb{E}_{\sim \xi_t} \left[ \|\gradftau\|\, \|\gradf\| \right]\geq \mathbb{E}_{\sim \xi_t} \left[ \|\gradf\|^{2} \right],
    \end{equation}
    since stale versions of the gradient should be larger, in magnitude, than recent versions.

    Combining everything together we rewrite our initial inequality as:

    \begin{flalign*}
        & \Delta_t=\mathbb{E}_\Omega \left[f\left ( \xvect_{t+1} \right ) -\fxt\right]\leq -\etat\rho\mu \mathbb{E}_{\sim \xi_t} \left[ \|\gradf\|^{2} \right]\\&+\frac{\etat^2 L}{2}\left(\mathbb{E}_{\sim \xi_t} \left[ \left \| \gradftau \right \|^{2} \right ] + \sigmasq\right).
    \end{flalign*}

    To derive a convergence bound it is necessary to sum all the increments over $t$ from $0$ to $T$:
    \begin{flalign*}
        &\sumT\Delta_t\leq \sumT\left(-\etat\rho\mu \mathbb{E}_{\sim \xi_t} \left[ \|\gradf\|^{2} \right]+\right.\\&\left.\frac{\etat^2 L}{2}\left(\mathbb{E}_{\sim \xi_t} \left[ \left \| \gradftau \right \|^{2} \right ] + \sigmasq\right)\right)\\
        &\leq \left(\sumT-\etat\rho\mu \mathbb{E}_{\sim \xi_t} \left[ \|\gradf\|^{2} \right]+\right.\\&\left.\frac{\etat^2 L}{2}\left(\mathbb{E}_{\sim \xi_t} \left[ \left \| \gradf \right \|^{2} \right ]+ \sigmasq\right)\right)+C=
        \\&\left(\sumT\left(-\etat\rho\mu+\frac{\etat^2 L}{2}\right) \mathbb{E}_{\sim \xi_t} \left[ \|\gradf\|^{2} \right]+ \frac{\etat^2 L}{2}\sigmasq\right)+C,
    \end{flalign*}
    where for the second inequality we used the result of \Cref{bounding}.

    We further manipulate the result by noticing that:
    $\sumT\Delta_t=\mathbb{E}_{\Omega}\left [f(\xvect_{t+1})-f(\xvect_{0})\right]$.
    Moreover since $
    \Lambda =\mathbb{E}_{\Omega}\left [f \left ( \xvect_{0} \right )\right]- \infop\limits_{\xvect} \fx,
    $
    it is easy to show that $\Lambda \geq \mathbb{E}_{\Omega}\left [f(\xvect_{0})-f(\xvect_{t+1})\right]=-\sumT\Delta_t$.
    We combine the bounds together as:
    \begin{flalign*}
        &-\Lambda \leq\sumT\Delta_t\leq \left(\sumT\left(-\etat\rho\mu+\frac{\etat^2 L}{2}\right) \mathbb{E}_{\sim \xi_t} \left[ \|\gradf\|^{2} \right]\right.\\&\left.+ \frac{\etat^2 L}{2}\sigmasq\right)+C.
    \end{flalign*}
    Then:
    \begin{flalign*}
        &-\Lambda -C\leq \left(\sumT\left(-\etat\rho\mu+\frac{\etat^2 L}{2}\right) \mathbb{E}_{\sim \xi_t} \left[ \|\gradf\|^{2} \right]\right.\\&\left.+ \frac{\etat^2 L}{2}\sigmasq\right),
    \end{flalign*}
    and finally:
    \begin{flalign*}
        &\left(\sumT\left(\etat\rho\mu-\frac{\etat^2 L}{2}\right) \mathbb{E}\left[ \|\gradf\|^{2} \right]\right)\\&\leq \left(\sumT\left(\frac{\etat^2 L}{2}\sigmasq\right)\right)+\Lambda+C.
    \end{flalign*}
    We conclude that:
    \begin{flalign*}
        \minop\limits_{0\leq t\leq T}\egrad\leq \frac{\left(\sumT\left(\frac{\etat^2 L}{2}\sigmasq\right)\right)+\Lambda+C}{\sumT\left(\etat\rho\mu-\frac{\etat^2 L}{2}\right)}.
    \end{flalign*}
    We then proceed by prooving the simple  \cref{corollary:convergence} of \Cref{theor:convergence}. By choosing a suitable constant learning rate $\etat=\eta=\frac{\rho\mu}{L\sqrt{T}}$ we can rewrite the inequality as
    \begin{flalign*}
        \minop\limits_{0\leq t\leq T}\egrad\leq \frac{\left(T\left(\frac{\eta^2 L}{2}\sigmasq\right)\right)+\Lambda^{'}}{T\left(\eta\rho\mu-\frac{\eta^2 L}{2}\right)}.
    \end{flalign*}
    where for simplicity we have defined $\Lambda^{'}=\Lambda+C$.

    We then notice that $\eta\rho\mu=\frac{(\rho\mu)^2}{L\sqrt{T}} $ and that $\frac{\eta^2 L}{2}=\frac{(\rho\mu)^2}{2LT}$, consequently we can rewrite the upper bound as
    \begin{flalign*}
        &\frac{\frac{(\rho\mu)^2\sigmasq}{2L}+\Lambda^{'}}{\frac{(\rho\mu)^2\sqrt{T}}{L}-\frac{(\rho\mu)^2}{2L}}=\frac{\frac{(\rho\mu)^2\sigmasq}{2L}+\Lambda^{'}}{\frac{(\rho\mu)^2}{L}\left(\sqrt{T}-\frac{1}{2}\right)}.
    \end{flalign*}
    Since we are intersted in $\mathcal{O}$ convergence rate, we can safely assume that $\left(\sqrt{T}-\frac{1}{2}\right)\simeq \sqrt{T}$ and that thus
    \begin{flalign*}
        &\minop\limits_{0\leq t\leq T}\egrad\lesssim \frac{\frac{(\rho\mu)^2\sigmasq}{2L}+\Lambda^{'}}{\frac{(\rho\mu)^2}{L}\sqrt{T}}\\&=\left(\frac{\sigmasq}{2}+\frac{\Lambda^{'}L}{(\rho\mu)^2}\right)\frac{1}{\sqrt{T}}
    \end{flalign*}

    \section{The simulator and a note on staleness}\label{sec:sim_staleness}
\begin{figure*}[t!]
        \centering
        \resizebox{10.0cm}{!}{
        \input{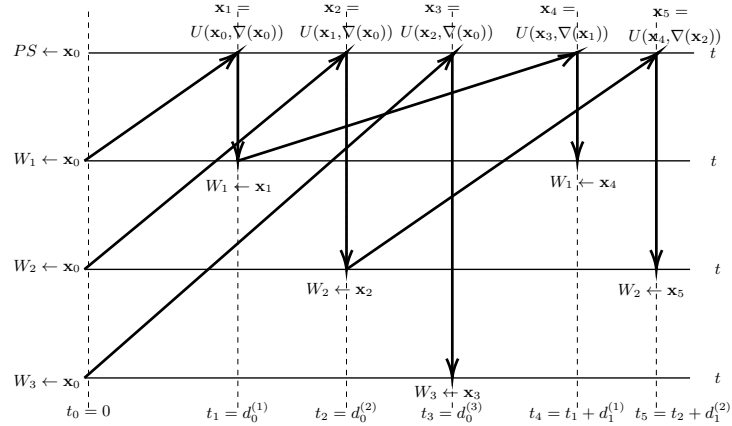}
        }
        \caption{Illustration of the distributed system operation. Example with one PS and 3 workers.}
        \label{fig:dsmsg}
    \end{figure*}

    In this Section we provide additional details to clarify the simulator structure, the definition of staleness and its generating process
    in our empirical validation.
    \Cref{fig:dsmsg} is an abstract representation of the distributed architecture we consider in this work.
    It consists of a time-line for each machine in the system: one for the \emph{parameter server} (PS), and one for each
    worker $W_i$ ($i=3$ in the Figure).

    In our system model, the PS accepts contributions from workers and updates the current model iterate according to
    the variant of the SGD algorithm it implements. We use the notation $\mathbf{x}_n=U(\mathbf{x}_{n-1},\nabla(\mathbf{x}_k))$ to indicate that the $n_{th}$ model version at PS is obtained by updating the $n-1_{th}$ version using a gradient computed with the $k_{th}$ version, according to the rules of the generic update algorithm $U(\cdot)$. In this case the staleness of the model update is equal  to $n-1-k$. It is assumed that as soon as the PS update the parameters the worker immediately receives the updated model, indicated with $W_i\leftarrow \mathbf{x}_n$. Each worker is going to transmit many different updates ($N_{updates}$), each of which will reach the PS after a random delay. We define the sequence of delays for the $i_{th}$ worker as $\{t_{\textsc{comm},r}^{(i)}\}_{r=0}^{N_{updates}}$. The random generation process will be shortly after explained.

    We are then ready to explain the example depicted in \Cref{fig:dsmsg}:
    \begin{enumerate}
        \item At time instant $t_0=0$ all workers and PS are initialized with model $\mathbf{x}_0$.
        \item At $t_1=t_{\textsc{comm},0}^{(1)}$ the PS receives the first gradient computed by $W_1$ using model $\mathbf{x}_0$. The PS updates the model ($\mathbf{x}_{1}=U(\mathbf{x}_{0},\nabla (\mathbf{x}_{0}))
      $) and send the update to $W_1$. In this case the staleness of the update is 0
      \item At $t_2=t_{\textsc{comm},0}^{(2)}$ the PS receives the first gradient computed by $W_2$ using model $\mathbf{x}_0$. The PS updates the model ($\mathbf{x}_{2}=U(\mathbf{x}_{1},\nabla (\mathbf{x}_{0}))
      $) and send the update to $W_2$. In this case the staleness is $1-0=1$.
      \item At $t_3=t_{\textsc{comm},0}^{(3)}$ the PS receives the first gradient computed by $W_3$ using model $\mathbf{x}_0$. The PS updates the model ($\mathbf{x}_{3}=U(\mathbf{x}_{2},\nabla (\mathbf{x}_{0}))
      $) and send the update to $W_3$. In this case the staleness is $3-1=2$.
    \item At $t_4=t_1+t_{\textsc{comm},1}^{(1)}$ the PS receives the second gradient computed by $W_1$ using model $\mathbf{x}_1$. The PS updates the model ($\mathbf{x}_{4}=U(\mathbf{x}_{3},\nabla (\mathbf{x}_{1}))
      $) and send the update to $W_1$. In this case the staleness is $2-0=2$.
     \item At $t_5=t_2+t_{\textsc{comm},1}^{(2)}$ the PS receives the second gradient computed by $W_2$ using model $\mathbf{x}_2$. The PS updates the model ($\mathbf{x}_{5}=U(\mathbf{x}_{4},\nabla (\mathbf{x}_{2}))
      $) and send the update to $W_2$. In this case the staleness is $4-2=2$.
      \item $\dots$
    \end{enumerate}
    
    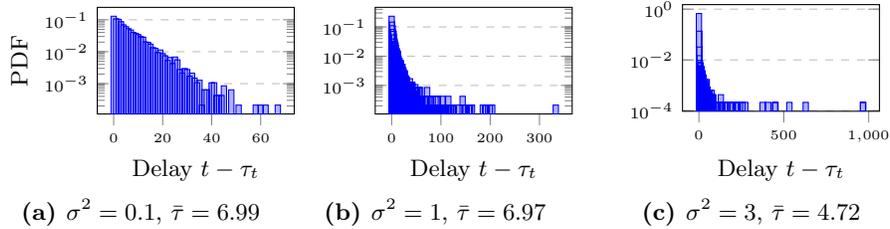
\begin{figure*}[!t]
        \hspace{-1.2em}
        \subfloat[$\sigmasq=0.1$, $\avtau=6.99$]{%
        \small
        \pgfplotsset{width=4.2cm}
        \pgfplotsset{height=3.0cm}
        \pgfplotsset{every x tick label/.append style={font=\fontsize{4}{4}\selectfont}}
        \pgfplotsset{every y tick label/.append style={font=\fontsize{6}{4}\selectfont}}
        \begin{tikzpicture}
    \begin{axis}[
    ymode=log,
    log origin=infty,
    xlabel={Delay $t-\tau_t$},
    ylabel={PDF},
    y tick label style={
    /pgf/number format/fixed,
    /pgf/number format/precision=1,
    },
    ybar,
    bar width=2pt,
    xtick pos=left,
    ymajorgrids=true,
    grid style=dashed,
    ]

\addplot plot coordinates
            {(0,0.128785) (1,0.106610) (2,0.103412) (3,0.086780) (4,0.069083) (5,0.060554) (6,0.052452) (7,0.047761) (8,0.040085) (9,0.037527) (10,0.034328) (11,0.029424) (12,0.028571) (13,0.020256) (14,0.019616) (15,0.017271) (16,0.013433) (17,0.012793) (18,0.009382) (19,0.011301) (20,0.008742) (21,0.007463) (22,0.005330) (23,0.006610) (24,0.004264) (25,0.006823) (26,0.005544) (27,0.002772) (28,0.002132) (29,0.002985) (30,0.002772) (31,0.001706) (32,0.002132) (33,0.001493) (34,0.001066) (35,0.001279) (36,0.000213) (37,0.000640) (38,0.000640) (40,0.001066) (41,0.001066) (42,0.000426) (43,0.000426) (44,0.000426) (45,0.000853) (48,0.000640) (51,0.000213) (58,0.000213) (60,0.000213) (62,0.000213) (67,0.000213)};

    \end{axis}
\end{tikzpicture}\label{fig:var01}
        }
         \hspace{-1.2em}
        \subfloat[$\sigmasq=1$, $\avtau=6.97$]{%
        \small
        \pgfplotsset{width=4.2cm}
        \pgfplotsset{height=3.0cm}
        \pgfplotsset{every x tick label/.append style={font=\fontsize{4}{4}\selectfont}}
        \pgfplotsset{every y tick label/.append style={font=\fontsize{6}{4}\selectfont}}
        \begin{tikzpicture}
    \begin{axis}[
    ymode=log,
    log origin=infty,
    xlabel={Delay $t-\tau_t$},
    y tick label style={
    /pgf/number format/fixed,
    /pgf/number format/precision=1,
    },
    ybar,
    bar width=2pt,
    xtick pos=left,
    ymajorgrids=true,
    grid style=dashed,
    ]

\addplot plot coordinates
            {(0,0.232196) (1,0.145629) (2,0.110235) (3,0.081663) (4,0.070576) (5,0.053305) (6,0.038806) (7,0.032409) (8,0.028358) (9,0.022388) (10,0.018550) (11,0.016844) (12,0.014072) (13,0.014925) (14,0.010021) (15,0.007036) (16,0.008529) (17,0.006610) (18,0.005970) (19,0.007036) (20,0.004904) (21,0.003412) (22,0.005330) (23,0.004051) (24,0.003838) (25,0.002559) (26,0.003198) (27,0.002985) (28,0.002132) (29,0.002132) (30,0.001279) (31,0.001919) (32,0.000853) (33,0.001279) (34,0.000640) (35,0.001279) (36,0.001279) (37,0.001706) (38,0.001066) (40,0.001279) (41,0.000640) (42,0.001279) (43,0.000640) (44,0.000640) (45,0.000640) (48,0.001066)
            (49,0.000640) (50,0.000853) (51,0.000213) (52,0.000853)
            (53,0.000426) (54,0.000213) (55,0.000853) (56,0.000213) (57,0.000213) (58,0.000853) (60,0.000426) (61,0.000426)
            (62,0.000213) (64,0.000213) (65,0.000213) (66,0.000213) (67,0.000213) (68,0.000426) (69,0.000213) (70,0.000213)
            (72,0.000853) (73,0.000213) (74,0.000213) (75,0.000426) (76,0.000213) (79,0.000213) (80,0.000426) (81,0.000213)
            (84,0.000213) (85,0.000213) (86,0.000213) (87,0.000213) (88,0.000213) (89,0.000426) (91,0.000213) (92,0.000213)
            (93,0.000213) (94,0.000426) (97,0.000213) (100,0.000213)
            (103,0.000426) (104,0.000213) (91,0.000213) (105,0.000213)
            (107,0.000213) (110,0.000213) (114,0.000213) (115,0.000426)
            (116,0.000213) (119,0.000213) (121,0.000213) (124,0.000213)
            (125,0.000213) (126,0.000213) (137,0.000213) (144,0.000426)
            (149,0.000213) (150,0.000213) (157,0.000213) (158,0.000213)
            (159,0.000213) (163,0.000213) (185,0.000213) (190,0.000213)
            (191,0.000213) (199,0.000213) (204,0.000213) (333,0.000213)};

    \end{axis}
\end{tikzpicture}\label{fig:var1}
        }
        \subfloat[$\sigmasq=3$, $\avtau=4.72$]{%
        \small
        \pgfplotsset{width=4.2cm}
        \pgfplotsset{height=3.0cm}
        \pgfplotsset{every x tick label/.append style={font=\fontsize{4}{4}\selectfont}}
        \pgfplotsset{every y tick label/.append style={font=\fontsize{6}{4}\selectfont}}
        \begin{tikzpicture}
    \begin{axis}[
    ymode=log,
    log origin=infty,
    xlabel={Delay $t-\tau_t$},
    y tick label style={
    /pgf/number format/fixed,
    /pgf/number format/precision=1,
    },
    ybar,
    bar width=2pt,
    xtick pos=left,
    ymajorgrids=true,
    grid style=dashed,
    ]

\addplot plot coordinates
            {(0,0.673561) (1,0.135181) (2,0.032196) (3,0.01472) (4,0.007676) (5,0.008316) (6,0.009595) (7,0.006397) (8,0.007463) (9,0.005330) (10,0.004904) (11,0.005544) (12,0.004478) (13,0.004264) (14,0.004051) (15,0.005544) (16,0.003198) (17,0.003412) (18,0.003412) (19,0.003838) (20,0.001706) (21,0.003412) (22,0.002985) (25,0.001066) (24,0.002559) (25,0.001066) (26,0.002559) (27,0.002345) (28,0.001279) (29,0.001706) (30,0.000853) (31,0.001066) (32,0.001279) (33,0.000213) (34,0.001493) (35,0.001279) (36,0.001066) (37,0.001066) (38,0.000426) (39,0.000853) (40,0.001066) (41,0.001066) (42,0.000853) (43,0.001493) (44,0.000853) (45,0.001066) (46,0.001706) (47,0.000640) (48,0.000640) (49,0.000640) (51,0.000853) (52,0.000426) (53,0.000640) (54,0.000641) (55,0.000426) (56,0.000640) (57,0.000426) (58,0.000213) (59,0.000213) (60,0.000640) (61,0.000640) (62,0.000213) (64,0.000213) (65,0.000213) (66,0.000426) (67,0.000213) (68,0.000426) (70,0.000213) (72,0.000213) (74,0.000640) (76,0.000426) (77,0.000213) (78,0.000426) (80,0.000213) (81,0.000213) (83,0.000213) (85,0.000426) (87,0.000213) (96,0.000213) (98,0.000213) (113,0.000213) (129,0.000426)
            (136,0.000213) (138,0.000213) (154,0.000213) (164,0.000213)
            (185,0.000213) (200,0.000213) (216,0.000213) (223,0.000213)
            (229,0.000213) (256,0.000213) (273,0.000213) (375,0.000213)
            (408,0.000213) (448,0.000213) (451,0.000213) (533,0.000213)
            (629,0.000213) (966,0.000213) (967,0.000213)};

    \end{axis}
\end{tikzpicture}\label{fig:var3}
        }
        \caption{Delay distributions of a simulation run with \lenet on \mnist, in a distributed setting with 8 workers. For each worker we generate a network delay according to an exponential distribution with rate $\lambda$. We sample $\lambda$ from a log-normal distribution with mean 0 and variance $\sigmasq$. For each configuration, we also report the resulting average staleness $\avtau$.
        }
        \label{fig:staleness_distributions_app}
    \end{figure*}
    
    Each communication delay is generated according to an exponential probability distribution with rate $\lambda$.
    To simulate network hetereogenity each worker has its own (fixed throughout the simulation) rate $\lambda_i$.
    These rates are extracted independently for each worker according to a lognormal distribution, i.e. $\ln(\lambda_i)\sim \mathcal{N}(0,\sigmasq)$ where $\sigmasq$ is a user defined parameter that can be used to change statistical configurations of the system.
    We use \Cref{fig:staleness_distributions_app} to illustrate the delay distribution for an entire simulation, that is, from
    the first model iterate, until the end of the training phase.
    Every time the PS receives a contribution from a worker, it increments the count in the bin corresponding to the
    staleness of the stochastic gradient message.
    In the Figure we vary $\sigmasq$ to obtain different delay distributions.
    Using a larger variance induces very large delay values (long tail in the distribution).
    However, if we observe the average staleness $\avtau$, we notice it decreases, because the mass of the distribution is
    concentrated on the left, which in practice means the majority of workers experience small delays.
    
    From an implementation point of view, our work is divided in two tasks: the simulation of the delays and the simulation of the distributed architecture. The delays simulation is simply a generation of a sequence of random variables according to a given distribution, while to simulate the distributed system it is sufficient allocate in memory $N_{worker}+1$ versions of the model (workers and PS) and update them according to the order of arrival of messages. It is important to notice that the simulator is transparent to the underling mechanisms for computing gradients and updates.
    In this work, we use the \textsc{PyTorch-1.4} library to describe models, and how they are trained.
    The interface between the high and low level layers allows exchanging model iterates, and gradients.
    Note that we use automatic differentiation from \textsc{PyTorch-1.4} to compute gradients.
    
    Overall, the above software design allows to: 1) easily introduce new models to study, as they can be readily imported
    from legacy \textsc{PyTorch-1.4} code; 2) easily implement variants of stochastic optimization algorithms; 3) experiment
    with a variety of system configurations, including heterogeneity, various communication patterns, and various
    staleness distributions.
    
    \section{Train loss results}
    
    \begin{figure}[!h]
     \begin{center}
        \centering
        \small
        \pgfplotsset{width=6.0cm}
        \pgfplotsset{height=4.0cm}
        \pgfplotsset{every x tick label/.append style={font=\fontsize{4}{4}\selectfont}}
        \pgfplotsset{every y tick label/.append style={font=\fontsize{6}{4}\selectfont}}
        \input{./figures/train_loss_s0.1.tex}
    \end{center}
    
    \caption{Train loss for \lenet on \mnist, in a system with 8 workers and $\sigmasq=0.1$}
    \label{fig:tr_loss_mnist_var0.1}
    \end{figure}
    
    \begin{figure}[!h]
     \begin{center}
        \centering
        \small
        \pgfplotsset{width=6.0cm}
        \pgfplotsset{height=4.0cm}
        \pgfplotsset{every x tick label/.append style={font=\fontsize{4}{4}\selectfont}}
        \pgfplotsset{every y tick label/.append style={font=\fontsize{6}{4}\selectfont}}
        \input{./figures/train_loss_s1.0.tex}
    \end{center}
    
    \caption{Train loss for \lenet on \mnist, in a system with 8 workers and $\sigmasq=1.0$}
    \label{fig:tr_loss_mnist_var1.0}
    \end{figure}
    
    \begin{figure}[!h]
     \begin{center}
        \centering
        \small
        \pgfplotsset{width=6.0cm}
        \pgfplotsset{height=4.0cm}
        \pgfplotsset{every x tick label/.append style={font=\fontsize{4}{4}\selectfont}}
        \pgfplotsset{every y tick label/.append style={font=\fontsize{6}{4}\selectfont}}
        \input{./figures/train_loss_s3.0.tex}
    \end{center}
    
    \caption{Train loss for \lenet on \mnist, in a system with 8 workers and $\sigmasq=3.0$}
    \label{fig:tr_loss_mnist_var3.0}
    \end{figure}
    
    \begin{figure}[!h]
     \begin{center}
        \centering
        \small
        \pgfplotsset{width=6.0cm}
        \pgfplotsset{height=4.0cm}
        \pgfplotsset{every x tick label/.append style={font=\fontsize{4}{4}\selectfont}}
        \pgfplotsset{every y tick label/.append style={font=\fontsize{6}{4}\selectfont}}
        \input{./figures/train_loss_resnet.tex}
    \end{center}
    
    \caption{Train loss for \resnet on \cifar, in a system with 8 workers and $\sigmasq=0.1$}
    \label{fig:tr_loss_resnet}
    \end{figure}

    \section{Detailed experimental settings}\label{sec:experiment_details}
    
        \begin{table}[!h]
        \begin{tabular}{|c|c|c|c|}
            \hline
            $\eta$ & \ssgdm   & \ssgdm  & \vanilla  \\
            \hline
            \lenet & 0.01 & 0.01 & 0.01 \\
            \hline
            \resnet& 0.1 & 0.1 & 0.1 \\
            \hline
            \resnet{} - 32 workers & 0.005 & 0.005 & 0.005 \\
            \hline
        \end{tabular}
        \caption{Learning rate parameters. }
        \label{tab:eta}
    \end{table}

    \begin{table}[!h]
        \begin{tabular}{|c|c|c|c|}
            \hline
            momentum & \ssgdm   & \ssgdm  & \vanilla \\
            \hline
            \lenet & 0.5 & 0.5 & 0.5 \\
            \hline
            \resnet& 0.9 & 0.9 & 0.9 \\
            \hline
        \end{tabular}
        \caption{Momentum parameters.}
        \label{tab:mom}
    \end{table}

    \paragraph{General parameters}

    \begin{itemize}
        \item Number of simulation runs per experimental setting: 5, with best parameters
        \item Number of workers = $\{1,2,4,8,16,32,64,128\}$
        \item $t_{\text{\textsc{comm}}} \sim \mathit{Exp}\left(\lambda \right)$
        \item $\ln(\lambda) \sim \mathcal{N}(0,\sigmasq)$, $\sigmasq=\{0.1,1,3\}$
    \end{itemize}


    \paragraph{Models.}

    All model parameters we used, including learning rates and momentum, are specified
    in \Cref{tab:eta} and \Cref{tab:mom}. Additional details are as follows.

    \begin{itemize}
        \item Training epochs = 5 (\lenet), 161  (\resnet)
        \item Training mini-batch size = 64 (\lenet), 128 (\resnet)
        \item Testing mini-batch size = 64 (\lenet), 128 (\resnet)
    \end{itemize}

    The \cnn models are implemented in \textsc{PyTorch-1.x}.
    \begin{itemize}
        \item \lenet: architecture, $d=61706$
        \item \resnet: architecture, $d=590426$
    \end{itemize}
\end{document}